\documentclass[10pt,twocolumn,letterpaper]{article}

\usepackage[pagenumbers]{cvpr}      

\usepackage{times}
\usepackage{epsfig}
\usepackage{graphicx}
\usepackage{amsmath}
\usepackage{amssymb}
\usepackage{booktabs}
\usepackage{multirow}
\usepackage{makecell}
\usepackage{color}
\usepackage{listings}
\usepackage{multicol}
\usepackage{caption}
\usepackage{subcaption}
\usepackage{enumitem}
\usepackage{algorithm}
\usepackage{algorithmic}
\usepackage{mdframed}
\usepackage{tablefootnote}

\usepackage[switch]{lineno}

\usepackage{soul}

\usepackage[dvipsnames]{xcolor}

\definecolor{cvprblue}{rgb}{0.21,0.49,0.74}
\usepackage[pagebackref,breaklinks,colorlinks,citecolor=cvprblue]{hyperref}

\def\paperID{12608} 
\def\confName{CVPR}
\def\confYear{2024}

\title{Instruct-ReID: A Multi-purpose Person Re-identification Task with Instructions}

\author{\textbf{Weizhen He}$^{1}$\footnotemark[2] \quad \textbf{Yiheng Deng}$^1$ \quad \textbf{Shixiang Tang}$^{2,3}$\footnotemark[1] \quad \textbf{Qihao Chen}$^4$ \\
\textbf{Qingsong Xie}$^5$ \quad \textbf{Yizhou Wang}$^2$ \quad \textbf{Lei Bai}$^2$ \quad \textbf{Feng Zhu}$^3$ \quad \textbf{Rui Zhao}$^{3,6}$ \\
\textbf{Wanli Ouyang}$^{2}$ \quad \textbf{Donglian Qi}$^1$ \quad \textbf{Yunfeng Yan}$^{1}\footnotemark[1]$\\
$^1$Zhejiang University \quad $^2$Shanghai AI Laboratory  \\ $^3$SenseTime Research \quad  $^4$Liaoning Technical University \quad $^5$Shanghai Jiao Tong University \quad \\
$^6$Qing Yuan Research Institute, Shanghai Jiao Tong University \\
hewz@zju.edu.cn, yvonnech@zju.edu.cn}

\begin{document}
\maketitle

\begin{abstract}
\vspace{-1em}
Human intelligence can retrieve any person according to both visual and language descriptions. However, the current computer vision community studies specific person re-identification (ReID) tasks in different scenarios separately, which limits the applications in the real world. This paper strives to resolve this problem by proposing a new instruct-ReID task that requires the model to retrieve images according to the given image or language instructions. Our instruct-ReID is a more general ReID setting, where existing \textbf{6} ReID tasks can be viewed as special cases by designing different instructions. We propose a large-scale OmniReID benchmark and an adaptive triplet loss as a baseline method to facilitate research in this new setting. Experimental results show that the proposed multi-purpose ReID model, trained on our OmniReID benchmark without fine-tuning,  can improve  \textbf{+0.5\%}, \textbf{+0.6\%}, \textbf{+7.7\%} mAP on Market1501, MSMT17, CUHK03 for traditional ReID, \textbf{+6.4\%}, \textbf{+7.1\%}, \textbf{+11.2\%} mAP on PRCC, VC-Clothes, LTCC for clothes-changing ReID, \textbf{+11.7\%} mAP on COCAS+ real2 for clothes template based clothes-changing ReID when using only RGB images, \textbf{+24.9\%} mAP on COCAS+ real2 for our newly defined language-instructed ReID, \textbf{+4.3\%} on LLCM for visible-infrared ReID, \textbf{+2.6\%} on CUHK-PEDES for text-to-image ReID. The datasets, the model, and code are available at \href{https://github.com/hwz-zju/Instruct-ReID}{\textcolor[RGB]{255 20 147}{https://github.com/hwz-zju/Instruct-ReID}}.
\end{abstract}
\section{Introduction}
\label{sec:intro}

Identifying individuals exhibiting significant appearance variations from multi-model descriptions is a fundamental aspect of human intelligence with broad applications~\cite{nanay2018multimodal,lin2023multimodality,lu2023cross,Zhao_2021_WACV}. To imbue our machine with this capability, person re-identification (ReID)~\cite{zheng2017unlabeled,zheng2016person,Ci_2023_CVPR} has been introduced to retrieve images of the target person from a vast repository of surveillance videos or images across locations and time~\cite{gheissari2006person,Wang_2022_CVPR}. Recently, significant advancements have been made in developing precise and efficient ReID algorithms and establishing benchmarks covering various scenarios, such as traditional ReID~\cite{zheng2015scalable,chen2017person,wei2018person,Zheng_2021_ICCV}, clothes-changing ReID (CC-ReID)~\cite{huang2021clothing,gu2022clothes,jin2022cloth,shu2021semantic,hong2021fine}, clothes template based clothes-changing ReID (CTCC-ReID)~\cite{yu2020cocas,li2022cocas+}, visible-infrared ReID (VI-ReID)~\cite{liu2022learning,yang2022augmented,yang2022learning,zhang2022fmcnet} and text-to-image ReID (T2I-ReID)~\cite{bai2023rasa,chen2018improving,zheng2020dual,li2017person,yin2017adversarial}. However, focusing solely on one specific scenario possesses inherent limitations. For instance, customers must deploy distinct models to retrieve persons according to the query information, which significantly increases the cost for model training and results in inconvenience for the application.  To facilitate real-world deployment, there is a pressing need to devise one generic framework capable of re-identifying individuals across all scenarios mentioned above.

\begin{figure*}
  \centering
  \includegraphics[width=0.87\linewidth]{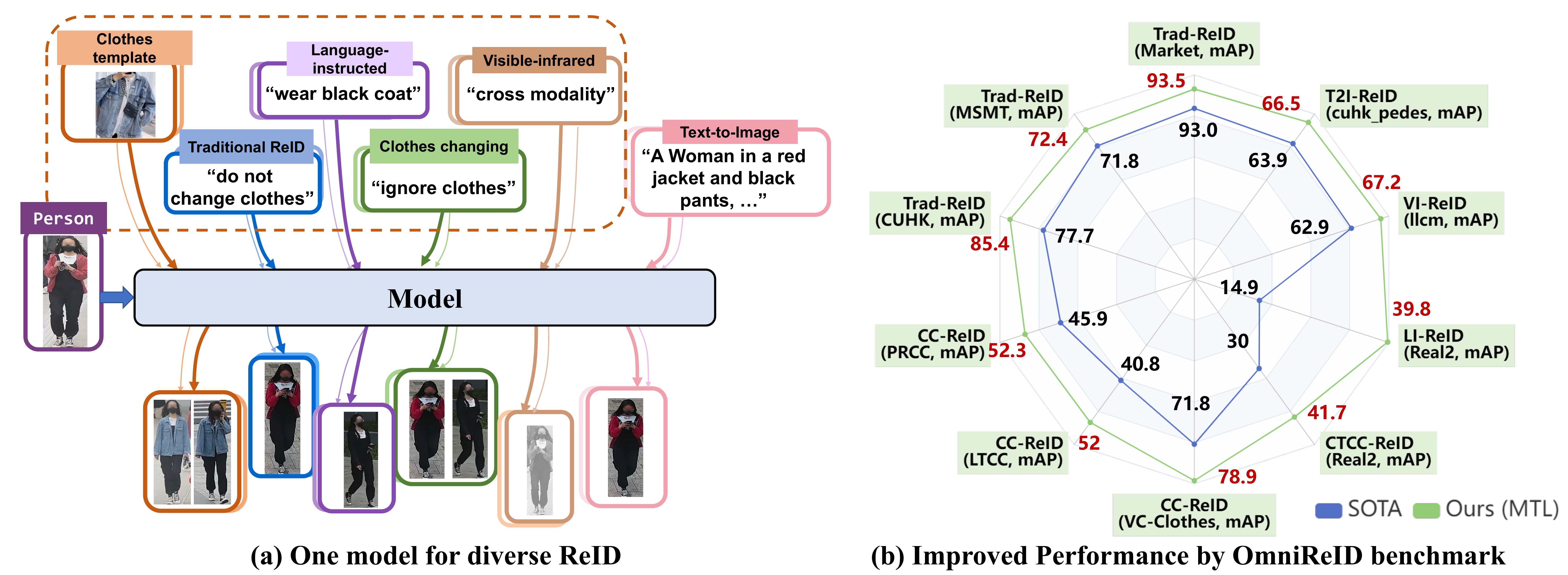}
  \caption{\textbf{(a)} We proposed a new instruct-ReID task that unites various ReID tasks. \emph{Traditional ReID:} The instruction may be “Do not change clothes". \emph{Clothes-changing ReID:} The instruction may be “Ignore clothes". \emph{Clothes template based clothes-changing ReID:} The instruction is a cropped clothes image and the model should retrieve the same person wearing the provided clothing. \emph{Language-instructed ReID:} The instruction is several sentences describing pedestrian attributes. The model is required to retrieve the person described by the instruction. \emph{Visible-Infrared ReID:} The instruction can be “Cross modality". \emph{Text-to-image ReID:} The model retrieves images according to the description sentence. \textbf{(b)} Our proposed method improves the performance of various person ReID tasks by a unified retrieval model. }
  \label{fig:teaser}
\end{figure*}
\setlength{\belowcaptionskip}{-0.3cm}

This paper proposes a new multi-purpose instruct-ReID task where existing \textbf{6} ReID settings can be formulated as its special cases. Specifically, the instruct-ReID uses query images and multi-model instructions as the model inputs, requiring the model to retrieve the same identity images from the gallery following the instructions. Using language or image instructions, models trained on the instruct-ReID task can be specialized to tackle diverse ReID tasks (Fig.~\ref{fig:teaser}a). For example, the clothes-changing ReID can be viewed as using the instruction “Ignore clothes" to retrieve. As another example, clothes template-based clothes-changing ReID can utilize a clothes template image as instruction. The proposed instruct-ReID task offers three significant advantages: \ul{easy deployment, improved performance, and easy extension to new ReID tasks}. First, it enables cost-effective and convenient deployment in real-world applications. Unlike existing ReID approaches limited to specific tasks, instruct-ReID allows for utilizing a single model for various ReID scenarios, which is more practical in real applications. Second, as all existing ReID tasks can be considered special cases of instruct-ReID, we can unify the training sets of these tasks to exploit the benefits of more data and diverse annotations across various tasks -- leading to enhanced performance. Third, instruct-ReID addresses a new language-instructed ReID task which requires the model to retrieve persons following language instructions. This setting is very practical in real applications for it enables customers to retrieve images they are particularly interested in through language descriptions. For example, customers can retrieve a woman wearing a black coat using one of her pictures and language instructions “Wear black coat".

To facilitate research in instruct-ReID, we introduce a new benchmark called OmniReID derived from 12 datasets \footnote{The discussion on ethical risks is provided in supplementary materials.} of 6 different ReID tasks. The OmniReID benchmark exhibits two appealing characteristics. First, it emphasizes \emph{\textbf{diversity}} by incorporating images from various domains, including surveillance scenarios and synthetic games. The diversity ensures that the trained models are robust and can effectively handle ReID tasks in various real-world scenarios. Second, the OmniReID benchmark achieves \emph{\textbf{comprehensiveness}} by offering evaluation datasets to assess various ReID tasks comprehensively, which facilitates evaluating the generalization ability of diverse ReID methods. 

Based on OmniReID, we design a method with the proposed adaptive triplet loss for instruct-ReID. The typical triplet loss~\cite{schroff2015facenet} only defines positive/negative pairs by identities, failing to align with instructions. Therefore, we propose a novel adaptive triplet loss to learn a metric space that preserves identity and instruction similarities. Specifically, we design an adaptive margin between two query-instruction pairs based on instruction similarities to pull features with similar instructions close and push features with different instructions apart. This loss incorporates the instruction information into the features representation and optimize the model to learn a metric space where similar instructions lead to closer features of query-instance pairs.

In summary, the contributions of this paper are three folds. (1) We propose a new instruct-ReID task, where existing traditional ReID, clothes-changing ReID, clothes template based clothes-changing ReID, visible-infrared ReID, text-to-image ReID andlanguage-instructed ReID can be viewed as special cases. (2) To facilitate research on instruct-ReID, we establish a large-scale and comprehensive OmniReID benchmark consisting 12 publicly available datasets. (3) We propose an adaptive triplet loss to supervise the feature distance of two query-instruction pairs to consider identity and instruction alignments. Our method consistently improves previous models on 10 datasets of 6 ReID tasks. For example, our method improves \textbf{+7.7\%}, \textbf{+0.6\%}, \textbf{+0.5\%} mAP on CUHK03, MSMT17, Market1501 for traditional ReID, \textbf{+6.4\%}, \textbf{+11.2\%}, \textbf{+7.1\%} mAP on PRCC, LTCC, VC-Clothes for clothes-changing ReID when using RGB images only, \textbf{+11.7\%} mAP on COCAS+ real2 for clothes template based clothes changing ReID, \textbf{+4.3\%} mAP on LLCM for visible-infrared ReID, \textbf{+2.6\%} mAP on CUHK-PEDES for text-to-image ReID, \textbf{+24.9\%} mAP on COCAS+ real2 for the new language-instructed ReID. 

\vspace{-0.5em}
\section{Related work}
\vspace{-0.5em}
\noindent \textbf{Person Re-identification.}
Person re-identification aims to retrieve the same images of the same identity with the given query from the gallery set. To support the ReID task on all-weather application, various tasks are conducted on the scenarios with changing environments, perspectives, and poses~\cite{ye2021deep,chen2017person,zheng2016person,chen2018improving,zheng2020dual,zhou2023adaptive,su2020adapting}. Traditional ReID mainly focuses on dealing with indoor/outdoor problems when the target person wears the same clothes. To extend the application scenarios, clothes-changing ReID (CC-ReID)~\cite{yu2020cocas} and clothes template based clothes-changing ReID (CTCC-ReID)~\cite{li2022cocas+} are proposed. While CC-ReID forces the model to learn clothes-invariant features~\cite{huang2021clothing,gu2022clothes}, CCTC-ReID further extracts clothes template features~\cite{li2022cocas+} to retrieve the image of the person wearing template clothes. To capture person's information under low-light environments, visible-infrared person ReID (VI-ReID) methods~\cite{zhang2022fmcnet,yang2022augmented,yang2022learning,zhang2022fmcnet} retrieve the visible (infrared) images according to the corresponding infrared (visible) images. In the absence of the query image, GNA-RNN~\cite{li2017person} introduced the text-to-image ReID (T2I-ReID) task, which aims at retrieving the person from the textual description. However, existing researches focus on a single scenario, making it difficult to address the demands of cross-scenario tasks. In this paper, we introduce a new Instruct-ReID task, which can be viewed as a superset of the existing ReID tasks by incorporating instruction information into identification.

\noindent  \textbf{Instruction Tuning.}
Instruction Tuning was first proposed to enable language models to execute specific tasks by following natural language instructions. Instruction-tuned models, \emph{e.g.}, FLAN-T5~\cite{chung2022scaling}, InstuctGPT~\cite{ouyang2022training}/ChatGPT~\cite{openai_chatgpt}, UPT~\cite{he2023unsupervised} can effectively prompt the ability on zero- and few-shot transfer tasks. A few works borrowed the idea from language to vision. Flamingo~\cite{alayrac2022flamingo}, BLIP-2~\cite{li2023blip}, and KOSMOS-1~\cite{huang2023language} learning with image-text pairs also show promising generalization on visual understanding tasks. While these methods aim to generate convincing language responses following the image or language instructions, we focus on retrieving the correct person following the given instructions by tuning a vision transformer.

\noindent \textbf{Multi-model Retrieval.}
Multi-model retrieval is widely used to align information from multiple modalities and improve the performance of applications. In multi-model retrieval, unimodal encoders always encode different modalities for retrieval tasks. For instance, CLIP~\cite{radford2021learning}, VideoCLIP~\cite{xu2021videoclip}, COOT~\cite{ging2020coot} and MMV~\cite{alayrac2020self} utilize contrastive learning for pre-training. Other techniques like HERO~\cite{li2020hero}, Clipbert~\cite{lei2021less}, Vlm~\cite{xu2021vlm}, TAM~\cite{nie2024triplet} and MVLT~\cite{ji2023masked} focus on merging different modalities for retrieval tasks to learn a generic representation. 
Although there have been numerous studies on multi-modal retrieval, most are concentrated on language-vision pretraining or video retrieval, leaving the potential of multi-modal retrieval for person ReID largely unexplored. This paper aims to investigate this underexplored area to retrieve anyone with information extracted from multiple modalities.

\vspace{-0.5em}
\section{OmniReID Benchmark}
\vspace{-0.5em}
To facilitate research on Instruct-ReID, we propose the OmniReID benchmark including a large-scale pretraining dataset based on 12 publicly available datasets with visual and language annotations. The comparison with the existing ReID benchmark is illustrated in Tab.~\ref{OmniReID benchmark}. 

\begin{table}
    \caption{Comparison of training subsets of different ReID datasets.}
    \label{OmniReID benchmark}
    \centering
    \footnotesize
    \resizebox{\linewidth}{!}{
    \begin{tabular}{l|ccc} 
    \toprule
    dataset    & image~ & ID   & domain                    \\ 
    \midrule
    MSMT17~\cite{wei2018person}     & 30,248  & 1,041 & indoor/outdoor            \\
    Market1501~\cite{zheng2015person} & 12,936  & 751  & outdoor                   \\
    PRCC~\cite{yang2019person}       & 17,896  & 150  & indoor                    \\
    COCAS+ Real1~\cite{li2022cocas+}      & 34,469  & 2,800 & indoor/outdoor            \\ 
    LLCM~\cite{zhang2023diverse}      & 30,921  & 713 & indoor/outdoor            \\ 
    LaST~\cite{shu2021large}      & 71,248  & 5,000 & indoor/outdoor \\
    MALS~\cite{yang2023towards}      & 1,510,330  & 1,510,330 & synthesis \\
    LUPerson-T~\cite{shao2023unified}      & 957,606  & - & indoor/outdoor \\
    \hline\hline
    OmniReID   & 4,973,044 & 328,604 & indoor/outdoor/synthetic  \\
    \bottomrule
    \end{tabular}
    }
\end{table}

\noindent \textbf{Protocols.}
To enable all-purpose person ReID, we collect massive public datasets from various domains and use their training subset as our training subset, including Market1501~\cite{zheng2015person}, MSMT17~\cite{wei2018person}, CUHK03~\cite{li2014deepreid} for traditional ReID, PRCC~\cite{yang2019person}, VC-Clothes~\cite{wan2020person}, LTCC~\cite{qian2020long} for clothes-changing ReID, LLCM~\cite{zhang2023diverse} for visible-infrared ReID, CUHK-PEDES~\cite{li2017person}, SYNTH-PEDES~\cite{zuo2023plip} for text-to-image ReID, COCAS+ Real1~\cite{zhang2023diverse} for clothes template based clothes-changing ReID and language-instructed ReID, forming 4,973,044 images and 328,604 identities. To fairly compare our method with state-of-the-art methods, the trained models are evaluated on LTCC, PRCC, VC-Clthoes, COCAS+ Real2, LLCM, CUHK-PEDES, Market1501, MSMT17, and CUHK03 test subsets without finetuning. We search the target images with query images and the instructions. Since a query image has multiple target images in the gallery set and CMC (Cumulative Matching Characteristic) curve only reflects the retrieval precision of most similar target images, we also adopt mAP (mean Average Precision) to reflect the overall ranking performance w.r.t. all target images. We present all dataset statistics of OmniReID in the supplementary materials.

\begin{figure}
  \centering
  \includegraphics[width=\linewidth]
  {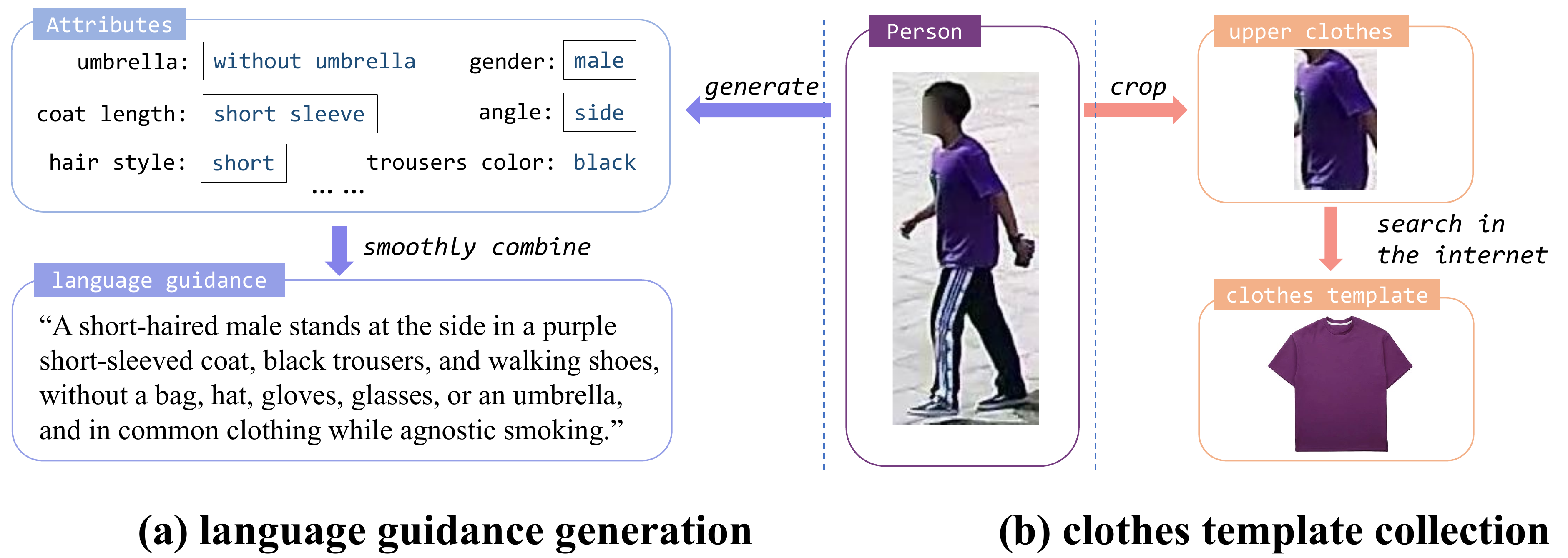}
  \caption{\textbf{(a)} We generate attributes for a person and then transform attributes into sentences by a large language model. \textbf{(b)} We crop upper clothes and search them online for clothes templates.}
  \label{fig:attr}
\end{figure}

\noindent \textbf{Language Annotation Generation.}
To generate language instruction for language-instructed ReID, we annotate COCAS+ Real1 and COCAS+ Real2 with language description labels. Similar to Text-to-Image ReID datasets, language annotations in OmniReID are several sentences that describe the visual appearance of pedestrians. We divide our annotation process into \emph{pedestrian attribute generation} and \emph{attribution-to-language transformation}. \\
\underline{\emph{Pedestrian Attribute Generation.}} To obtain a comprehensive and varied description of an individual, we employ an extensive collection of attribute words describing a wide range of human visual characteristics. The collection contains 20 attributes and 92 specific representation words, including full-body clothing, hair color, hairstyle, gender, posture, and accessories such as umbrellas or satchels. Professional annotators manually label all the pedestrian attributes. We provide a practical illustration with the attribute collection in Fig.~\ref{fig:attr}(a). By utilizing instructions on the well-defined attribute combination, models can further enhance their ability to identify the target person. \\
\underline{\emph{Attribute-to-Language Transformation.}}
Compared with discrete attribute words, language is more natural for consumers. To this end, we transform these attributes into multiple sentences using the Alpaca-LoRA~\cite{zhang2023llama} large language model. Specifically, we ask the Alpaca-LoRA with the following sentences: “Generate sentences to describe a person. The above sentences should contain all the attribute information I gave you in the following." 
The generated annotations are carefully checked and corrected manually to ensure the correctness of the language instructions. All detailed pedestrian attributes and more language annotations are presented in the supplementary materials.

\begin{figure*}
  \centering
  \includegraphics[width=0.95\linewidth]{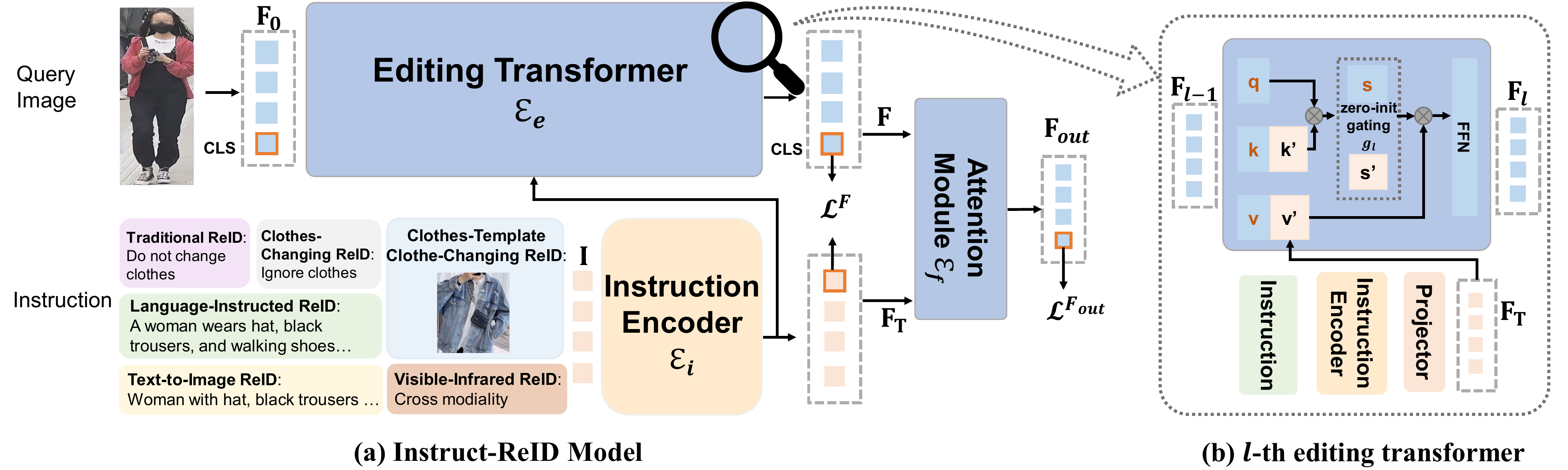}
  \caption{The overall architecture of the proposed method. The instruction is fed into the instruction encoder to extract instruction features (a). The features are then propagated into the editing transformer (b) to capture instruction-edited features. We exploit adaptive triplet loss and identification loss to train the network. For the testing stage, we use the CLS token for retrieval. }
  \label{fig:method}
\end{figure*}
\noindent \textbf{Visual Annotation Generation.} Visual annotations are images that describe the characteristics of pedestrians. In this paper, we select clothes as the visual annotations because they are viewed as the most significant visual characteristics of pedestrians. To get high-quality visual annotations, we first crop the upper clothes from the source images (Clothes Copping) and search on the internet to get the most corresponding clothes-template images (Clothes-template Crawling) as visual annotations. Since each person wears the same clothes in traditional ReID datasets, we annotate the clothes-changing LTCC dataset where each person wears multiple clothes to ease the burden of annotations. Fig.~\ref{fig:attr} (b) shows the detailed process. \\
\underline{\emph{Clothes Cropping.}} We use a human parsing model SCHP~\cite{li2020self} to generate the segmentation mask of the upper clothes and then crop the corresponding rectangle upper clothes patches from the original images. These bounding boxes of upper clothes are then manually validated. \\
\underline{\emph{Clothes-template Crawling.}} Given all cropped upper clothes from images in OmniReID datasets, we crawl the templates of these clothes from shopping websites\footnote{https://world.taobao.com/, https://www.17qcc.com/}. The top 40 matching clothes templates are downloaded when we crawl each cropped upper clothes. The one with the highest image quality is manually selected.

\section{IRM: Instruct ReID Model}

 As shown in Fig.~\ref{fig:method} (a), the proposed Instruct ReID model consists of four parts: instruction generation (Sec.~\ref{sec:instruct_generation}) for various ReID tasks, an editing transformer $\mathcal{E}_{e}$ (Sec.~\ref{sec:edit_transformer}) and an instruction encoder $\mathcal{E}_{i}$ which is a visual language model, and a cross-model attention module $\mathcal{E}_{f}$. 
Given an instruction $\mathbf{T}$ associated with a query image $\mathbf{I}$, we obtain instruction features $\mathbf{F}_\mathbf{T}$ using the instruction encoder $\mathcal{E}i$. These extracted instruction features $\mathbf{F}_{T}$, along with query image tokens, are fed into the designed editing transformer $\mathcal{E}_e$  to obtain features $\mathbf{F}$ edited by instructions. Furthermore, we introduce an attention module $\mathcal{E}_f$ to efficiently combine features of the query image and instruction through the attention mechanism in the stacked transformer layers. Finally, the overall loss function is introduced (Sec.~\ref{sec:overall_loss}) for all ReID tasks, which includes our newly proposed adaptive triplet loss $\mathcal{L}_{atri}$ (Sec.~\ref{sec:adaptive_triplet_loss}) instead of traditional triplet loss.

\subsection{Instruction Generation} \label{sec:instruct_generation}

In our proposed instruct-ReID task, the model must retrieve the images that describe the same person following the instructions. By designing different instructions, our instruct-ReID can be specialized to existing ReID tasks, \emph{i.e.,} traditional ReID, clothes-changing ReID, clothes template based clothes-changing ReID, visual-infrared ReID, text-to-image ReID, and language-instructed ReID. We show the current instructions and leave the exploration of better instructions for instruct-ReID to future research.

\noindent \textbf{Traditional ReID:} Following Instruct-BLIP~\cite{instructblip},  we generate \underline{20} instructions\footnote{See all available instruction sentences in supplementary materials.} from GPT-4 and randomly select one, \emph{e.g.,} “Do not change clothes.", when training. The model is expected to retrieve images of the same person without altering image attributes, such as clothing. 
\begin{mdframed}[backgroundcolor=gray!10]
\textbf{\#\#\# Query image}:\{Query image\} \\ \quad \textbf{\#\#\# Instruction}: “Do not change clothes."$^3$ \\
\textbf{\#\#\# Target image}:\{Output\}
\end{mdframed}
\textbf{Clothes-changing ReID:} Simiar to Traditional ReID, the instruction is the sentence chosen from a collection of \underline{20} GPT-4 generated sentences$^3$, \emph{e.g.,} “With clothes changed". The model should retrieve images of the same person even when wearing different outfits. 
\begin{mdframed}[backgroundcolor=gray!10]
\textbf{\#\#\# Query image}:\{Query image\} \\ \quad \textbf{\#\#\# Instruction}: “With clothes changed."$^3$ \\
\textbf{\#\#\# Target image}:\{Output\}
\end{mdframed}
\textbf{Clothes template based clothes-changing ReID:} The instruction is a clothes template for a query image while a cropped clothes image for a target image. The model should retrieve images of the same person wearing the provided clothes. We provide more examples for training and test in the supplementary materials.
\begin{mdframed}[backgroundcolor=gray!10]
\textbf{\#\#\# Query image}:\{Query image\} \\ \quad \textbf{\#\#\# Instruction}:\{Any clothes template image\}\\
\textbf{\#\#\# Target image}:\{Output\}
\end{mdframed}
\textbf{Visual-Infrared ReID:} The instruction is the sentence chosen from a collection of \underline{20} GPT-4 generated sentences$^3$, \emph{e.g.,} “Retrieve cross modality image". The model should retrieve visible (infrared) images of the same person according to the corresponding infrared (visible) images. 
\begin{mdframed}[backgroundcolor=gray!10]
\textbf{\#\#\# Query image}:\{Query image\} \\ \quad \textbf{\#\#\# Instruction}: “Retrieve cross modality image."$^3$ \\
\textbf{\#\#\# Target image}:\{Output\}
\end{mdframed}
\textbf{Text-to-Image ReID:} The instruction is the describing sentences, and both images and text are fed into IRM during the training process. While in the inference stage, the image features and instruction features are extracted separately. \footnote{Image and instruction features are extracted separately in test stage.} 
\begin{mdframed}[backgroundcolor=gray!10]
\textbf{\#\#\# Image}:\{Image\}$^4$ \\ \quad \textbf{\#\#\# Instruction}:\{Sentences describing pedestrians\}$^4$\\
\textbf{\#\#\# Target image}:\{Output\}
\end{mdframed}

\textbf{Language-instructed ReID:} The instruction is several sentences describing pedestrian attributes. We randomly select the description languages from the person images in gallery and provide to query images as instruction. The model is required to retrieve images of the same person described in the provided sentences. We provide more examples for training and test in the supplementary materials.
\begin{mdframed}[backgroundcolor=gray!10]
\textbf{\#\#\# Query image}:\{Query image\} \\ \quad \textbf{\#\#\# Instruction}:\{Sentences describing pedestrians\}\\
\textbf{\#\#\# Target image}:\{Output\}
\end{mdframed}

\subsection{Editing Transformer} \label{sec:edit_transformer} 
The editing transformer consists of $L$ zero-init transformer layers $\mathcal{E}_e = \{\mathcal{F}_1, \mathcal{F}_2, ..., \mathcal{F}_L\}$, which can leverage the instruction to edit the feature of query images. Given the $l$-th zero-init transformer, the output feature $\mathbf{F}_l$ can be formulated as 
\begin{equation} \small
    \mathbf{F}_l = \mathcal{F}_l(\mathbf{F}_{l-1}, \mathbf{F}_{\mathbf{T}}),
\end{equation}
where $\mathbf{F}_\mathbf{T} = \mathcal{E}_i(\mathbf{I})$ is the instruction feature extracted by $\mathcal{E}_i$ and $\mathbf{F}_{l-1}$ is the output feature of $(l\!-\!1)$-th zero-init transformer layer. The initial input $\mathcal{F}_0$ of the first layer is defined as $\mathbf{F}_0 = [\mathbf{f}_0^{\text{CLS}}, \mathbf{f}_0^{1}, \mathbf{f}_0^{2}, ..., \mathbf{f}_0^{N}],$
where $\mathbf{f}^{\text{CLS}}_0$ is the [CLS] token,  $(\mathbf{f}_0^{1}, \mathbf{f}_0^{2}, ..., \mathbf{f}_0^{N})$ are the patch tokens of the query image and $N$ is the number of patches of the query image.

We show the detailed structure of each layer in the editing transformer in Fig.~\ref{fig:method} (b). Given the features $\mathbf{F}_{l-1}$ and instruction features $\mathbf{F}_{\mathbf{T}}$, the attention map $\mathbf{M}_l$ is defined as 
\begin{equation} \small
    \mathbf{M}_l = \left[\text{Softmax}(\mathbf{S}_l), g_l \times \text{Softmax}(\mathbf{S'}_l)\right],
\end{equation}
where $g_l$ is the gating parameters initialized by zero. Here, $\mathbf{S}_l$ is  the attention map between queries and keys of input features and $\mathbf{S}'_l$ is the attention map between queries of input features and keys of instruction features. Mathematically, 
\begin{equation} \small
    \mathbf{S}_l = \mathbf{Q}_l\mathbf{K}_l^\top/\sqrt{C}, \mathbf{S}_l' = \mathbf{Q}_l\mathbf{K'}_l^\top/\sqrt{C},
\end{equation}
where a linear projection derives queries and keys, \emph{i.e.,} $\mathbf{Q}_l\!=\!\text{Linear}_q(\mathbf{F}_{l-1})$, $\mathbf{K}_l\!=\!\text{Linear}_k(\mathbf{F}_{l-1})$ and $\mathbf{K}'_l\!=\!\text{Linear}_{k'}(\mathbf{F}_{\mathbf{T}})$, respectively. $C$ is the feature dimension of query features. Finally, we calculate the output of the $l$-th layer by 
\begin{equation} \small
    \mathbf{F}_l = \text{Linear}_o(\mathbf{M}_l\left[\mathbf{V}_l, \mathbf{V}'_l\right]),
\end{equation}
where $\text{Linear}_o$ is the feed-forward network after the attention layer in each transformer block, $\mathbf{V}_l$ and $\mathbf{V}'_l$ are the values calculated by $\mathbf{V}_l=\text{Linear}_v(\mathbf{F}_{l-1})$ and $\mathbf{V}'_l=\text{Linear}_{v'}(\mathbf{F}_{\mathbf{T}})$. We use the [CLS] token in the output feature of $L$-th transformer layer for computing losses and retrieval, \emph{i.e.,} $\mathbf{F} = \mathbf{f}_L^{\text{CLS}}$, where $\mathbf{F}_L=(\mathbf{f}_L^{\text{CLS}}, \mathbf{f}_L^{1}, \mathbf{f}_L^{2}, ..., \mathbf{f}_L^{N})$ and $N$ is patch number of query images.

\subsection{Adaptive Triplet Loss} \label{sec:adaptive_triplet_loss}

Unlike typical triplet loss that defines positive and negative samples solely based on identities, instruct-ReID requires distinguishing images with different instructions for the same identity. Intuitively, an adaptive margin should be set to push or pull samples based on the instruction difference. Let ($\mathbf{F}^{a}_{i}$, $\mathbf{F}^{r_{1}}_{i}$, $\mathbf{F}^{r_{2}}_{i}$) be the $i$-th triplet in the current mini-batch, where $\mathbf{F}^{a}_{i}$ is an anchor sample, $\mathbf{F}^{r_{1}}_{i}$ and $\mathbf{F}^{r_{2}}_{i}$ are reference samples. We propose an adaptive triplet loss as  
\begin{equation} \small
    \begin{split}
    \mathbf{\mathcal{L}}_{atri}=\frac{1}{N_{tri\dagger} } \sum_{i=1}^{N_{tri\dagger}}  \{ \mathbf{Sign}(\beta_1 - \beta_2)  [  d(\mathbf{F}^{a}_{i},\mathbf{F}^{r_{1}}_{i}) + \\ \left ( \beta_1 - \beta_2  \right )m - d(\mathbf{F}^{a}_{i},\mathbf{F}^{r_{2}}_{i})  ]  \}_{+} 
    \end{split}
\end{equation}
where $N_{tri\dagger}$ and $m$ denote the number of triplets and a hyper-parameter for the maximal margin, respectively. $d$ is a Euclidean distance function, \emph{i.e.}, $d(\mathbf{F}^{a}_{i},\mathbf{F}^{r}_{i})= \left |\left | \mathbf{F}^{a}_{i} - \mathbf{F}^{r}_{i} \right | \right |_{2} ^{2}$. $\beta_1$ and $\beta_2$ are relatednesses between the anchor image and the corresponding reference image that consider the identity consistency and instruction similarity for the adaptive margin. Mathematically, they are defined as
\begin{equation} \small
    \beta_j = \mathbb{I}\left(y_a=y_{r_j}\right) \mathbf{Cos}\left \langle \mathbf{F}_{\mathbf{T}}^{a}, \mathbf{F}_{\mathbf{T}}^{r_{j}} \right \rangle,
\end{equation}
where $y_a$ and $y_{r_j}$ are the identity labels of the anchor image and the reference image, $\mathbb{I}(\cdot)$ is the indicator function, and $j=\{1, 2\}$ denotes the index of reference samples.

\begin{figure}
    \centering
    \includegraphics[width=\linewidth]{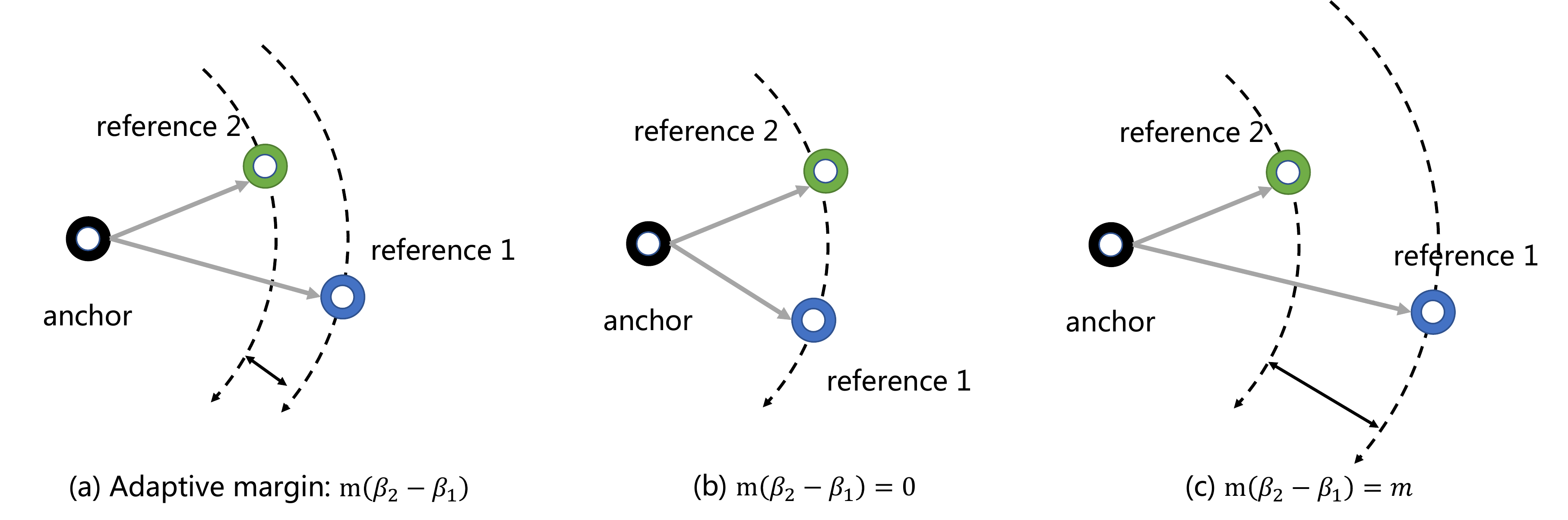}
    \caption{Illustration of adaptive triplet loss. Unlike the traditional triplet loss where the margin is fixed, the margin in our adaptive triplet loss is defined by the instruction similarity for the two query-instruction pairs that describe the same person. The features associated with similar instructions are pulled to be closer.}
    \label{fig:my_label}
\end{figure}

The concept of adaptive triplet loss is described by Fig.~\ref{fig:my_label}(a). We discuss the adaptive loss in two cases. First, as shown in Fig.~\ref{fig:my_label}(b), the margin is set to zero if the triplet has the same identity and the instructions of the two reference samples are equally similar to the instruction of the anchor sample. This makes the distances between the anchor point and the two reference points the same. Second, as shown in Fig.~\ref{fig:my_label}(c), when there is a significant difference in instruction similarities, the margin distance between the anchor and two references becomes closer to the maximum value $m$, forcing the model to learn discriminative features. Adaptive triplet loss makes features from the same person become distinctive based on the similarity of instructions, which helps to retrieve images that align the requirements of given instructions in the CTCC-ReID and LI-ReID tasks.

\subsection{Overall Loss Function} \label{sec:overall_loss} 

We impose an identification loss $\mathcal{L}_{id}$, which is the classification loss on identities, and an adaptive triplet loss $\mathcal{L}_{atri}$ on $\mathbf{F}$ and the final fusion features $\mathbf{F}_\mathbf{out}$ to supervise the model for training Trad-ReID, CC-ReID, VI-ReID, CTCC-ReID, and LI-ReID tasks. The overall loss is combined as
\begin{equation} \small
    \begin{split}
    \mathbf{\mathcal{L}}=\mathcal{L}_{atri}(F) + \mathcal{L}_{id}(F) + \mathcal{L}_{atri}(F_{out}) + \mathcal{L}_{id}(F_{out})
    \end{split}
\end{equation}
where $\mathbf{F}$ and $\mathbf{F}_{out}$ indicate the source of features used in calculating the loss.

For the T2I-ReID task, we adopt a contrastive loss $\mathcal{L}_{cl}$ to align the image features $\mathbf{F}$ and text features $\mathbf{F}_\mathbf{T}$ to enable text-based person retrieval. We also employ a binary classification loss $\mathcal{L}_{match}$ to learn whether an inputted image-text pair is positive or negative, defined as:
\begin{equation} \small
    \begin{split}
    \mathbf{\mathcal{L}}=\mathcal{L}_{cl}({F}) + \mathcal{L}_{match}({F_{out}})
    \end{split}
\end{equation}
where, $\mathcal{L}_{match}({F_{out}}) = \mathcal{C}_{e} (\hat{y},F_{out}) = \mathcal{C}_{e} [\hat{y},\mathcal{E}_f(\mathbf{F}, \mathbf{F_{T}})] $, respectively. $\mathcal{C}_{e}$ is a binary cross-entropy loss, $\hat{y}$ is a 2-dimension one-hot vector representing the ground-truth label \emph{i.e.,} $[0, 1]^\intercal$ for positive pair and $[1, 0]^\intercal$ for negative pair, formed by matching text features with corresponding image features before inputting into the attention module $\mathcal{E}_f$.
\vspace{-0.5em}
\section{Experiment}

\subsection{Experimental Setups}

\noindent \textbf{Training Settings.}
To enable all-purpose person ReID, we perform two training scenarios based on the built benchmark \textbf{OmniReID}: 1)
Single-task Learning (STL): Every task is trained and tested individually using the corresponding dataset.
2) Multi-task Learning (MTL): To acquire one unified model for all tasks, the model is optimized by joint training of the four ReID tasks with all the training datasets. The trained network is then evaluated for different tasks on various datasets. \ul{We provide the details of datasets used in STL and MTL in the supplementary materials.}

\noindent \textbf{Implementation Details.}
For the editing transformer, we use the plain ViT-Base with the ReID-specific pretraining~\cite{zhu2022pass}. The instruction encoder is ALBEF~\cite{li2021align}. All images are resized into 256$\times$128 for training and test. We use the AdamW optimizer with a base learning rate of 1e-5 and a weight decay of 5e-4 . We linearly warmup the learning rate from
1e-7 to 1e-5 for the first 1000 iterations. Random cropping, flipping, and erasing are used for data augmentation during training. For each training batch, we randomly select 32 identities with 4 image samples for each identity.

\begin{table}[t]
  \footnotesize
  \centering
  \renewcommand\arraystretch{1.2}
  \caption{The performance of Clothes-Changing ReID of our method and the state-of-the-art methods. Mean average precision (mAP) and Top1 are used to quantify the accuracy. \dag~denotes that the model is trained with multiple datasets. *~denotes that the model is pre-trained on 4 million images. }
    \begin{tabular}{l|cc|cc|cc}
    \hline
    \multicolumn{1}{l|}{\multirow{2}{*}{\textbf{Methods}}}&\multicolumn{2}{c|}{\textbf{LTCC}} &\multicolumn{2}{c|}{\textbf{PRCC}} &\multicolumn{2}{c}{\textbf{VC-Clothes}} \\
    \cline{2-7} & mAP & Top1 & mAP & Top1 & mAP & Top1\\
    \hline
    HACNN\citep{li2018harmonious} & 26.7 & 60.2 & - & 21.8 & - & -\\
    RGA-SC\citep{zhang2020relation} & 27.5 & 65.0 & - & 42.3 & 67.4 & 71.1\\
    PCB\citep{sun2018beyond} & 30.6 & 65.1 & 38.7 & 41.8 & 62.2 & 62.0\\
    IANet\citep{hou2019interaction} & 31.0 & 63.7 & 45.9 & 46.3 & - & -\\
    CAL\citep{gu2022clothes} & 40.8 & 74.2 & - & - & - & -\\
    TransReID\citep{he2021transreid} & - & - & - & 44.2 & 71.8 & 72.0\\
    \hline
    IRM (STL) & 46.7 & 66.7 & 46.0 & 48.1 & \textbf{80.1} & \textbf{90.1} \\
    IRM (MTL)\dag & \textbf{52.0} & \textbf{75.8} & \textbf{52.3} & \textbf{54.2} & 78.9 & 89.7 \\
    \hline
    \end{tabular}%
  \label{tab:CC-ReID}%
\end{table}%

\subsection{Experimental Results}
\noindent \textbf{Clothes-Changing ReID (CC-ReID).}
As shown in Tab~\ref{tab:CC-ReID}, IRM outperforms all state-of-the-art methods on LTCC, PRCC and VC-Clothes, showing that the model can effectively extract clothes-invariant features following the instructions, \emph{e.g.,} ``Ignore clothes''. Specifically, on STL, IRM improves CAL~\cite{gu2022clothes}, TransReID~\cite{he2021transreid} by \textbf{+5.9\%} mAP and \textbf{+8.3\%} mAP on LTCC and VC-Clothes, respectively. On MTL, IRM further improves the performance on LTCC to \textbf{52.0\%} mAP and reaches a new state-of-the-art on PRCC with \textbf{52.3\%} mAP. Other methods like ACID~\cite{yang2023win}, CCFA~\cite{han2023clothing}, AIM~\cite{yang2023good} and DCR-ReID~\cite{cui2023dcr} provide results of 384$\times$192 image size, we present more results based on this resolution of IRM in supplementary. While multi-task learning leads to slightly lower performance than single-task learning on VC-Clothes, our method still achieves a higher Top-1 than TransReID. We conjecture that this drop is due to the domain gap between VC-Clothes (Synthetic) and other datasets (Real) and leave it for future work.

\noindent \textbf{Clothes Template Based Clothes-Changing ReID (CTCC-ReID).}
Our method achieves desirable performance on the CTCC-ReID task in Tabel~\ref{tab:CT-ReID}, which shows that a fixed instruction encoder is enough for this tough task. Concretely, when only trained on COCAS+ Real1, IRM outperforms BC-Net~\cite{yu2020cocas} and DualBCT-Net~\cite{li2022cocas+}, both of which learn an independent clothes branch, by \textbf{+9.6\%} mAP and \textbf{+2.2\%} mAP, respectively. By integrating the knowledge on other instruct ReID tasks during multi-task learning, we are able to further improve the performance of IRM, achieving an mAP of \textbf{41.7\%} and pushing the performance limits on CTCC-ReID.

\begin{table}[t]\footnotesize
  \centering
  \renewcommand\arraystretch{1.2}
  \caption{Comparison with the state-of-the-art methods on visible-infrared ReID and text-to-image ReID. The VI-ReID setting is \textit{VIS-to-IR} and \textit{IR-to-VIS} in LLCM. \dag~denotes that the model is trained with multiple datasets.*~denotes that the model is trained under the same image shape as IRM, \emph{i.e.,} 256$\times$128.  }
    \begin{tabular}{l|cc|cccc}
    \hline
    \multicolumn{1}{l|}{\multirow{3}{*}{\textbf{Methods}}} &\multicolumn{2}{c|}{\textbf{T2I-ReID}}&\multicolumn{4}{c}{\textbf{VI-ReID: LLCM}} 
    \\
    \cline{2-7} 
    &\multicolumn{2}{c|}{\textbf{CUHK-PEDES}} & \multicolumn{2}{c}{\textbf{VIS-to-IR}} &\multicolumn{2}{c}{\textbf{IR-to-VIS}}  \\
    & mAP & Top1 & mAP & Top1 & mAP & Top1 \\
    \hline
    ALBEF~\cite{li2021align} & 56.7 & 60.3&- & -&-&- \\
    CAJ~\cite{ye2021channel} & - & - & 59.8 & 56.5 & 56.6 & 48.8\\
    MMN~\cite{zhang2021towards} & - & - & 62.7 & 59.9 & 58.9 & 52.5\\
    DEEN~\cite{zhang2023diverse} & - & - & 65.8 & 62.5 & 62.9 & 54.9\\
    SAF~\cite{li2022learning} & 58.6& 64.1& - & -&-&-\\
    PSLD~\cite{han2021text}& 60.1 & 64.1&-&-&-&- \\
    RaSa*~\cite{bai2023rasa}& 63.9 & 69.6 &-&-&-&- \\
    \hline
    IRM (STL) & 65.3 & 72.8 & 66.6 & 66.2 & 64.5 & 64.9 \\
    IRM (MTL)\dag & \textbf{66.5} & \textbf{74.2} & \textbf{67.5} & \textbf{66.7} & \textbf{67.2} & \textbf{65.7} \\
    \hline
    \end{tabular}%
  \label{tab:t2i_cross}%
\end{table}%

\begin{table*}[t]\small
  \centering
  \renewcommand\arraystretch{1.0}
  \caption{Performance comparison with the state-of-the-art methods on Clothes
  -Template Clothes-Changing ReID, Language-Instructed ReID, and Traditional ReID. \dag~denotes that the model is trained with multiple datasets. *~denotes that the model is pretrained on 4 million pedestrian images. \underline{The size of the input images used in the table is 256x128.} }
    \begin{tabular}{l|cc|cc|cccccc}
    \hline
    \multicolumn{1}{l|}{\multirow{3}{*}{\textbf{Methods}}} &\multicolumn{2}{c|}{\textbf{CTCC-ReID}}&\multicolumn{2}{c|}{\textbf{LI-ReID}} &\multicolumn{6}{c}{\textbf{Trad-ReID}} 
    \\
    \cline{2-11} 
    &\multicolumn{2}{c|}{\textbf{COCAS+ Real2}}&\multicolumn{2}{c|}{\textbf{COCAS+ Real2}} &\multicolumn{2}{c}{\textbf{Market1501}} &\multicolumn{2}{c}{\textbf{MSMT17}} &\multicolumn{2}{c}{\textbf{CUHK03}}  \\
    & mAP & Top1& mAP & Top1& mAP & Top1& mAP & Top1& mAP & Top1 \\
    \hline
    Baseline & - & -&14.9&31.6& -& -& -& -& -& -\\
    TransReID~\cite{he2021transreid} & 5.5 & 17.5&-&-&86.8 & 94.4 & 61.0 & 81.8 & - & -\\
    BC-Net~\cite{yu2020cocas} & 22.6 & 36.9&-&-& -& -& -& -& -& -\\
    DualBCT-Net~\cite{li2022cocas+} & 30.0 & 48.9&-&-& -& -& -& -& -& -\\
    SAN~\cite{jin2020semantics}&-&-&-&-&
    88.0 & 96.1 & - & - & 76.4 & 80.1
    \\
    HumanBench\dag~\cite{tang2023humanbench}&-&-&-&-&89.5 & - & 69.1 & - & 77.7 & -
    \\
    PASS*~\cite{zhu2022pass}&-&-&-&-&93.0 & 96.8 & 71.8 & 88.2 & - & - 
    \\
    \hline
    IRM (STL) & 32.2 & 54.8 & 30.7 & 60.8 & 92.3 & 96.2 & 71.9 & 86.2 & 83.3 & 86.5 \\
    IRM (MTL)\dag & \textbf{41.7} & \textbf{64.9} & \textbf{39.8} & \textbf{71.6} & \textbf{93.5} & \textbf{96.5} & \textbf{72.4} & \textbf{86.9} & \textbf{85.4} & \textbf{86.5} \\
    \hline
    \end{tabular}%
  \label{tab:CT-ReID}%
\end{table*}%

\noindent \textbf{Visible-Infrared ReID (VI-ReID).} We validate the performance of IRM on Visible-Infrared ReID datasets LLCM, which is a new and challenging low-light
cross-modality dataset and has a more significant number
of identities and images. From Tab.~\ref{tab:t2i_cross}, we can see that the results on the two test modes show that the proposed IRM achieves competitive performance against all other state-of-the-art methods. Specifically, for the VIS-to-IR mode on LLCM, IRM achieves \textbf{67.5\%} mAP and exceeds previous state-of-the-art methods like DEEN~\cite{zhang2023diverse} by \textbf{+1.7\%}. For the IR-to-VIS mode on LLCM, IRM achieves \textbf{65.7\%} Rank-1 accuracy
and \textbf{67.2\%} mAP, which is a new state-of-the-art result. The results validate the effectiveness of our method.

\noindent \textbf{Text-to-Image ReID (T2I-ReID).} As shown in Tab.~\ref{tab:t2i_cross}, IRM shows competitive performance with a mAP of \textbf{66.5\%} on the CUHK-PEDES~\cite{li2017person}, which is \textbf{+2.6\%}, \textbf{+6.4\%}, \textbf{+7.9\%} higher than previous methods  RaSa~\cite{bai2023rasa} (63.9\%), PSLD~\cite{han2021text} (60.1\%), SAF~\cite{li2022learning} (58.6\%). Because a few images in CUHK-PEDES training set are from the test sets of Market1501 and CUHK03, we filtered out duplicate images from the test sets during the multi-task learning (MTL) process. The testing was conducted on uniformly resized images with a resolution of 256$\times$128.

\begin{table*}[t]
  \centering
  \scriptsize
  \renewcommand\arraystretch{1.2}
  \caption{Ablation study. The performance comparison of our editing transformer and using ViT base transformer (w/o editing), and the comparisons with triplet loss (w/o $\mathcal{L}_{atri}$) and the proposed adaptive triplet loss in terms of mAP. \dag~denotes that the test mode is VIS-to-IR and \ddag~denotes IR-to-VIS mode on LLCM.}
    \begin{tabular}{l|c|c|c|cc|ccc|ccc|c}
    \hline
    \multicolumn{1}{l|}{\multirow{2}{*}{\textbf{Methods}}}&\textbf{CTCC-ReID}&\textbf{LI-ReID}&\textbf{T2I-ReID}&\multicolumn{2}{c|}{\textbf{VI-ReID}}&\multicolumn{3}{c|}{\textbf{CC-ReID}} &\multicolumn{3}{c|}{\textbf{Trad-ReID}} & \multirow{2}{*}{\textbf{Avg.}} \\
    \cline{2-12} & Real2 & Real2 & CUHK. & LLCM\dag & LLCM\ddag & LTCC & PRCC & VC-Clo. & Market1501 & MSMT17 & CUHK03 & \\
    \hline
    IRM (STL) & 32.2 & 30.7 & 65.3 & 66.6 & 64.5 & 46.7 & 46.0 & 80.1 & 92.3 & 71.9 & 83.3 & 61.8\\
    w/o editing & 32.6 & 30.5 & 65.7 & 66.2 & 65.1 & 46.2 & 45.7 & 77.8 & 92.5 & 71.4 & 83.2 & 61.5\\
    w/o $\mathcal{L}_{atri}$ & 31.5 & 30.2 & - & 66.6 & 64.5 & 46.7 & 46.0 & 80.1 & 92.3 & 71.9 & 83.3 & 61.3\\
    \hline
    IRM (MTL) & 41.7 & 39.8 & 66.5 & 67.5 & 67.2 & 52.0 & 52.3 & 78.9 & 93.5 & 72.4 & 85.4 & 65.1\\
   w/o editing & 41.0 & 38.8 & 66.1 & 67.3& 65.2 & 51.2 & 51.0 & 78.6 & 93.0 & 71.8 & 85.1 & 64.5\\
    w/o $\mathcal{L}_{atri}$ & 40.7 & 39.2 & 65.2 & 68.4 & 66.3 & 52.0 & 52.4 & 77.9 & 92.9 & 72.0 & 85.5 & 64.8\\
    \hline
    \end{tabular}%
  \label{tab:ablation study}%
\end{table*}%

\noindent \textbf{Language-Instructed ReID (LI-ReID).}
As a new setting in ReID, no previous works have been done to retrieve a person using several sentences as the instruction, therefore, we compare IRM with a straightforward baseline. In the baseline method, only a ViT-Base is trained on COCAS+ Real1 images without utilizing the information from language instruction, leading to poor person re-identification ability. As shown in Tab~\ref{tab:CT-ReID}, IRM improves the baseline by \textbf{+15.8\%} mAP, because IRM can integrate instruction information into identity features. With MTL, IRM achieves extra \textbf{+9.1\%} performance gain by using more images and general information in diverse ReID tasks.

\noindent \textbf{Traditional ReID (Trad-ReID).} 
IRM also shows its effectiveness on Trad-ReID tasks in Tab~\ref{tab:CT-ReID}. Specifically, when trained on a single dataset, compared with PASS~\cite{zhu2022pass}, IRM achieves comparable performance on Market1501, MSMT17 and \textbf{+6.9\%} performance gain on CUHK03 compared with SAN~\cite{jin2020semantics}. With multi-task training, IRM can outperform the recent multi-task pretraining HumanBench~\cite{tang2023humanbench} and self-supervised pretraining~\cite{zhu2022pass}. We do not compare with SOLDIER~\cite{Chen_2023_CVPR} because it only reports ReID results with the image size of 384x192 instead of 256x128 in our method. More importantly, SOLDIER primarily focuses on pretraining and is evaluated on traditional ReID tasks only, while the claimed contribution of IRM is to tackle multiple ReID tasks with one model.
\subsection{Ablation Study}
\textbf{Editing Transformer.}
To verify the effectiveness of the instruction integrating design in the editing transformer, we compare it with the traditional ViT base model, where the image features are extracted without fusing information from instruction features. Results in Tab~\ref{tab:ablation study} show that adopting the editing transformer leads to \textbf{-0.6\%} mAP performance drop in the MTL scenario. Consistent results can be observed in STL, indicating the effectiveness of the instruction integrating design in the editing transformer.

\noindent \textbf{Adaptive Triplet Loss.}
Tab~\ref{tab:ablation study} shows that adaptive triplet loss in MTL outperforms the traditional triplet loss by \textbf{+0.3\%} mAP on average, indicating that the proposed loss boosts the model to learn more discriminative features following different instructions. On STL, for CC-ReID and Trad-ReID, the instructions are fixed sentences leading to the same performance of adaptive/traditional triplet loss. However, in the case of CTCC-ReID and LI-ReID, where instructions vary among samples, using adaptive triplet loss brings about \textbf{+0.7\%}, \textbf{+0.5\%} mAP gain, which shows the effectiveness of adaptive triplet loss in learning both identity and instruction similarity.

\noindent \textbf{More Results.} We provide additional results based on different pre-trained models and visualizations of the retrieval results in the supplementary material.


\vspace{-1em}
\section{Conclusion}
\vspace{-0.5em}
This proposes one unified instruct-ReID task to jointly tackle existing traditional ReID, clothes-changing ReID, clothes template based clothes-changing ReID, language-instruct ReID, visual-infrared ReID, and text-to-image ReID tasks, which holds great potential in social surveillance. To tackle the instruct-ReID task, we build a large-scale and comprehensive OmniReID benchmark and a generic framework with an adaptive triplet loss. We hope our OmniReID can facilitate future works such as unified network structure design and multi-task learning methods on a broad variety of retrieval tasks.

\noindent \textbf{Acknowledgement}.This paper was supported by the National Natural Science Foundation of China (No.62127803), Key R\&D Project of Zhejiang Province (No.2022C01056).

{
    \small
    \bibliographystyle{ieeenat_fullname}
    \bibliography{main}
}


\section*{Appendix}

\appendix

\begin{table*}[ht]
  \centering
  \scriptsize
  \renewcommand\arraystretch{1.2}
  \caption{More results of IRM on different pre-trained models. We train and test using the default image resolution of 256$\times$128 (256), with (384) indicating an image resolution of 384$\times$192. \dag~denotes that the test mode is VIS-to-IR and \ddag~denotes IR-to-VIS mode on LLCM.}
    \resizebox{1.0\textwidth}{!}{\begin{tabular}{cc|c|c|c|cc|ccc|ccc}
    \toprule
    \multicolumn{2}{c|}{\multirow{2}{*}{\textbf{IRM}}}&\textbf{CTCC-ReID}&\textbf{LI-ReID}&\textbf{T2I-ReID}&\multicolumn{2}{c|}{\textbf{VI-ReID}}&\multicolumn{3}{c|}{\textbf{CC-ReID}} &\multicolumn{3}{c}{\textbf{Trad-ReID}} \\
    \cline{3-13} & & Real2 & Real2 & CUHK. & LLCM\dag & LLCM\ddag & LTCC & PRCC & VC-Clo. & Market1501 & MSMT17 & CUHK03  \\
    \midrule
    SOTA & - & 30.0~\cite{li2022cocas+} & 14.9 & 63.9~\cite{bai2023rasa} & 65.8~\cite{zhang2023diverse} & 62.9~\cite{zhang2023diverse} & 40.8~\citep{gu2022clothes} & 45.9~\citep{hou2019interaction} & 71.8~\citep{he2021transreid} & 93.0~\cite{zhu2022pass} & 71.8~\cite{zhu2022pass} & 77.7~\cite{tang2023humanbench}\\
    \midrule
    \multirow{2}{*}{DeiT~\cite{pmlr-v139-touvron21a}} & STL(256) & 31.5 & 30.1 & 64.2 & 65.1 & 62.3 & 46.3 & 43.1 & 82.1 & 87.9 & 67.9 & 77.5\\
    & MTL(256) & 40.2 & 38.7 & 65.7 & 66.2 & 62.9 & 53.2 & 46.8 & 76.3 & 90.0 & 68.9 & 78.7\\
    \midrule
    \multirow{2}{*}{ALBEF~\cite{li2021align}} & STL(256) & 30.6 & 28.7 & 67.7 & 61.2 & 58.8 & 40.2 & 42.1 & 67.1 & 78.9 & 60.3 & 69.2\\
    & MTL(256) & 37.2 & 34.5 & \textbf{68.3} & 62.1 & 60.1 & 41.2 & 44.1 & 60.5 & 79.7 & 62.1 & 69.9\\
    \midrule
    \multirow{2}{*}{HAP~\cite{yuan2023hap}} & STL(256) & 30.8 & 31.1 & 63.2 & 66.2 & 62.9 & 45.3 & 44.7 & 79.2 & 90.2 & 70.3 & 79.6\\
    & MTL(256) & 39.2 & 39.8 & 65.1 & 67.1 & 64.3 & 49.2 & 48.7 & 77.3 & 91.7 & 72.3 & 82.2\\
    \midrule
    \multirow{4}{*}{PASS~\cite{zhu2022pass}} & STL(256) & 32.2 & 30.7 & 65.3 & 66.6 & 64.5 & 46.7 & 46.0 & 80.1 & 92.3 & 71.9 & 83.3\\
    & MTL(256) & 41.7 & 39.8 & 66.5 & 68.5 & 67.2 & 52.0 & 52.3 & 78.9 & 93.5 & 72.4 & 85.4\\
    & STL(384) & 33.7 & 33.8 & 65.9 & 68.9 & 64.7 & 48.9 & 49.3 & \textbf{82.5} & 92.7 & 73.7 & 84.1\\
    & MTL(384) & \textbf{42.5} & \textbf{42.3} & 67.2 & \textbf{71.2} & \textbf{67.3} & \textbf{54.1} & \textbf{58.4} & 81.8 & \textbf{93.9} & \textbf{75.4} & \textbf{88.7}\\
    \bottomrule
    \end{tabular}}
  \label{tab:more results}%
\end{table*}%
\vspace{-1em}
\section{More results of IRM on different pre-trained models}
\vspace{-1em}
In this section, we provide more results about IRM on the publicly released pre-trained models \emph{i.e.,} DeiT~\cite{pmlr-v139-touvron21a}, {ALBEF~\cite{li2021align}, HAP~\cite{yuan2023hap}, and PASS~\cite{zhu2022pass} in Tab.~\ref{tab:more results}. We can see two conclusions from Tab.~\ref{tab:more results}. First, we can see models pre-trained on human-centric images, \emph{i.e.,} HAP~\cite{yuan2023hap} and PASS~\cite{zhu2022pass}, can naturally increase the performance of person-retrieval tasks. Second, although ALBEF~\cite{li2021align} exhibits lower performance on Trad-ReID and other image-based ReID tasks, it achieves the highest performance on the T2I-ReID task due to its vision-language pretraining knowledge. Additionally, we validate the effectiveness of IRM on 384$\times$192 (384) image resolution using the PASS pre-trained model. The results show a further improvement in performance compared to the 256$\times$128 results, which indicates that increasing the image size can contribute to achieving better retrieval results.
\vspace{-0.2cm}
\section{Visualizaton}
\vspace{-0.2cm}
We visualize the retrieval results of CTCC-ReID, LI-ReID, CC-ReID and Trad-ReID tasks in Fig.~\ref{fig:vis} and VI-ReID, T2I-ReID in Fig.~\ref{fig:vis_vi}. Given a query image, IRM not only retrieves the right person from the gallery but also finds specific target images of the person following the instruction. Concretely, for CTCC-ReID, IRM retrieves images of query persons wearing instructed clothes as shown in the first row. For LI-ReID, IRM effectively parses information from languages such as bag condition (\emph{e.g.}, row 2 person {\#1}) and clothes attribute (\emph{e.g.}, row 2 person {\#2,3}) to find the correct image. For CC-ReID, with the ``Ignore clothes'' instruction, our method focuses on biometric features and successfully retrieves the person in the case of changing clothes, \emph{e.g.} the  4th and 5th images of row 3 person {\#1}, which the clothes are different from the query image. For Trad-ReID shown in row 4, ``do not change clothes'' instructs IRM to pay attention to clothes, a main feature of a person's image. In this case, IRM retrieves images with the same clothes in the query image. For VI-ReID, we visualize the retrieval results for both VIS-to-IR and IR-to-VIS modes. IRM can retrieve the correct cross-modality images even in low visual conditions. For T2I-ReID, when there are images of different identities in the gallery but with representations consistent with the text description (\emph{e.g.}, row 1 3rd and row 2 person 4th images in Fig~\ref{fig:vis_vi}(c)), it may introduce some noise to IRM. However, IRM is still capable of indexing most of the correct results based on the given descriptions.

\begin{figure*}[t]
    \centering
    \includegraphics[width=\linewidth]{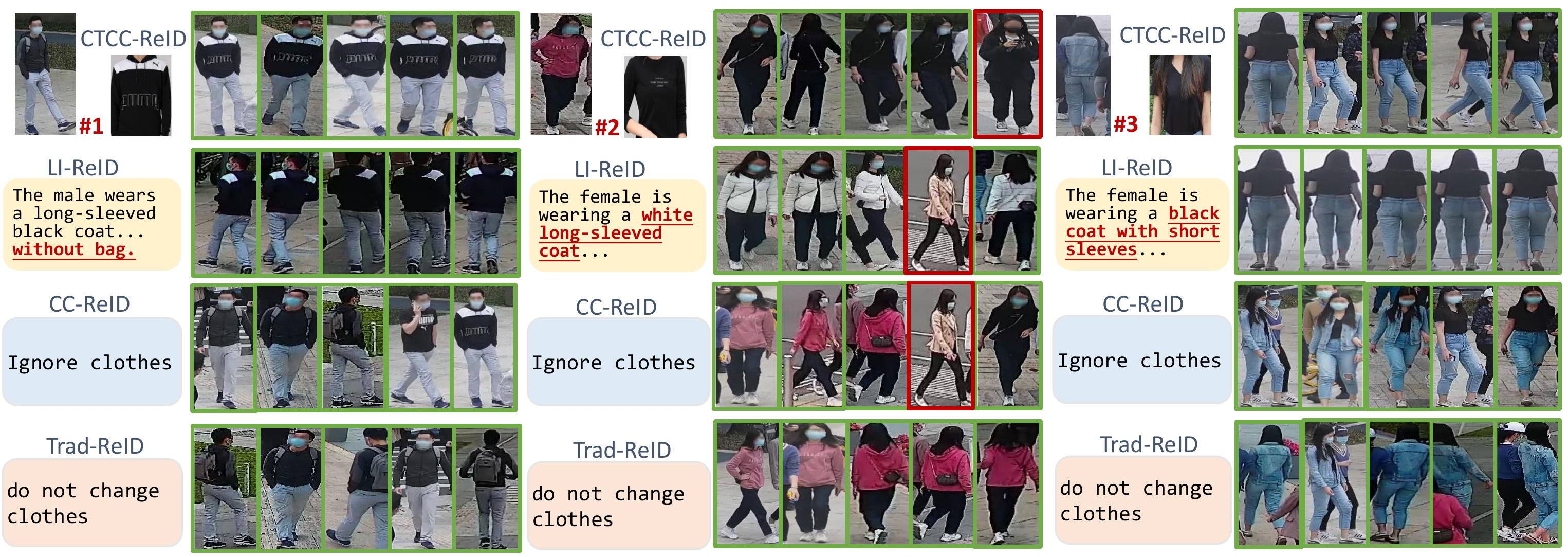}
    \caption{Illustration of all tasks retrieval results. We visualize the task-specific instructions on three people as examples. Green and red boxes mean true and false matches.}
    \label{fig:vis}
\end{figure*}

\begin{figure*}[t]
    \centering
    \includegraphics[width=\linewidth]{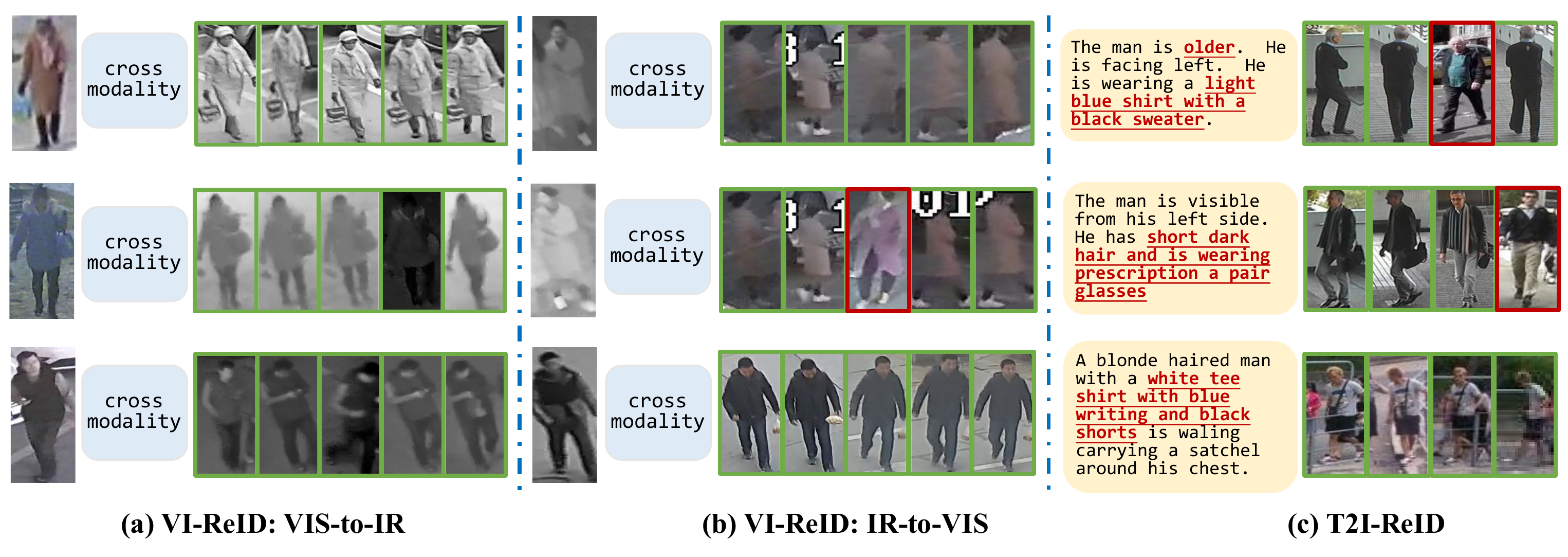}
    \caption{Illustration of VI-ReID and T2I-ReID tasks retrieval results. We visualize both VIS-to-IR and IR-to-VIS mode results on three people in the LLCM dataset. There are about four images for each person in  CUHK-PEDES, we visualize the top 4 results for T2I-ReID on the CUHK-PEDES dataset as examples. Green and red boxes mean true and false matches.}
    \label{fig:vis_vi}
\end{figure*}
\vspace{-0.2cm}
\section{Details of Language Annotation Generation}
\vspace{-0.2cm}
In this section, we provide more details about the generation of language annotation in OmniReID. To depict pedestrian images with more detailed descriptions, we first manually annotate OmniReID with local description attribute words, including clothes color, accessory, pose, etc. From a human perspective, language descriptions with sentences are much effective in describing person than simply listing attribute. Based on the attribute annotations we collect, we merge them into language annotations to provide a more comprehensive description of the person.

\subsection{Pedestrian Attribute Generation}

We annotate 20 attributes and 92 specific representation in words for OmniReID, as shown in Tab.~\ref{tab:attribute details}. The attributes are carefully selected considering a wide range of human visual characteristics from the datasets, including full-body clothing, hair color, hairstyle, gender, age, actions, posture, and accessories such as umbrellas or satchels. Though there might be more than one representation for attributes such as coat color and trousers color, only one representation corresponding to the image is selected for annotation. The attributes are annotated by professional annotators in the image level, thus the annotation file will contain more accurate and detailed description. For example, in Fig.~\ref{fig:attr_to_language_example} (a), although the two images are of the same identity, the difference of pedestrian angle is also annotated.

\begin{table*}[t]\scriptsize
  \centering
  \renewcommand\arraystretch{1.2}
  \caption{Details of attributes from OmniReID that describe person images. We select 20 attributes and 92 specific representation words considering a wide range of human visual appearances in detail.}
  \resizebox{0.96\textwidth}{!}{
    \begin{tabular}{@{}c|c}
    \toprule
    Attribute & representation in words \\
    \midrule
    \multicolumn{1}{m{2.2cm}|}{coat color} & \multicolumn{1}{m{10cm}}{"black coat", "blue coat", "gray coat", "green coat", "purple coat", "red coat", "white coat", "yellow coat"} \\ \hline
    \multicolumn{1}{m{2.2cm}|}{trousers color} & \multicolumn{1}{m{10cm}}{"black trousers", "blue trousers", "gray trousers", "green trousers", "purple trousers", "red trousers", "white trousers", "yellow trousers"} \\ \hline
    \multicolumn{1}{m{2.2cm}|}{coat length} & \multicolumn{1}{m{10cm}}{"agnostic length coat", "long sleeve coat", "short sleeve coat", " bareback coat"} \\ \hline
    \multicolumn{1}{m{2.2cm}|}{trousers length} & \multicolumn{1}{m{10cm}}{"shorts trousers", "skirt", "trousers"} \\ \hline
    \multicolumn{1}{m{2.2cm}|}{gender code} & \multicolumn{1}{m{10cm}}{"female", "agnostic gender", "male"} \\ \hline
    \multicolumn{1}{m{2.2cm}|}{glass style} & \multicolumn{1}{m{10cm}}{"without glasses", "with glasses", "with sunglasses"} \\ \hline
    \multicolumn{1}{m{2.2cm}|}{hair color} & \multicolumn{1}{m{10cm}}{"black hair", "agnostic color hair", "white hair", "yellow hair"} \\ \hline
    \multicolumn{1}{m{2.2cm}|}{hair style} & \multicolumn{1}{m{10cm}}{"bald hair", "agnostic style hair", "long hair", "short hair"} \\ \hline
    \multicolumn{1}{m{2.2cm}|}{bag style} & \multicolumn{1}{m{10cm}}{"backpack", "hand bag", "shoulder bag", "waist pack", "trolley", "agnostic style bag", "without bag"} \\ \hline
    \multicolumn{1}{m{2.2cm}|}{cap style} & \multicolumn{1}{m{10cm}}{"with hat", "without hat"} \\ \hline
    \multicolumn{1}{m{2.2cm}|}{shoes color} & \multicolumn{1}{m{10cm}}{"black shoes", "blue shoes", "gray shoes", "green shoes", "purple shoes", "red shoes", "white shoes", "yellow shoes"} \\ \hline
    \multicolumn{1}{m{2.2cm}|}{shoes style} & \multicolumn{1}{m{10cm}}{"boots", "leather shoes", "sandal", "walking shoes"} \\ \hline
    \multicolumn{1}{m{2.2cm}|}{age} & \multicolumn{1}{m{10cm}}{"adult", "child", "old"} \\ \hline
    \multicolumn{1}{m{2.2cm}|}{person angle} & \multicolumn{1}{m{10cm}}{"back", "front", "side"} \\ \hline
    \multicolumn{1}{m{2.2cm}|}{pose} & \multicolumn{1}{m{10cm}}{"lie", "pose agnostic", "sit", "stand", "stoop"} \\ \hline
    \multicolumn{1}{m{2.2cm}|}{coat style} & \multicolumn{1}{m{10cm}}{"business suit", "agnostic style coat", "dress", "jacket", "long coat", "shirt", "sweater", "t-shirt"} \\ \hline
    \multicolumn{1}{m{2.2cm}|}{glove} & \multicolumn{1}{m{10cm}}{"with glove", "agnostic glove", "without glove"} \\ \hline
    \multicolumn{1}{m{2.2cm}|}{smoking} & \multicolumn{1}{m{10cm}}{"smoking", "agnostic smoking", "without smoking"} \\ \hline
    \multicolumn{1}{m{2.2cm}|}{umbrella} & \multicolumn{1}{m{10cm}}{"with umbrella", "without umbrella"} \\ \hline
    \multicolumn{1}{m{2.2cm}|}{uniform} & \multicolumn{1}{m{10cm}}{"chef uniform", "common clothing", "firefighter uniform", "medical uniform", "office uniform", "agnostic uniform", "worker uniform"} \\ 
    \bottomrule
  \end{tabular}}
  \label{tab:attribute details}
\end{table*}

\subsection{Attribute-to-Language Transformation}
We provide some example from our OmniReID that provides sentence descriptions of individuals in images. Compared with discrete attribute words, language is more natural for consumers. To this end, we transform these attributes into multiple sentences using the Alpaca-LoRA large language model. Specifically, we ask the Alpaca-LoRA with the following sentences: "Generate sentences to describe a person. The above sentences should contain all the attribute information I gave you in the following." Annotators carefully check the generated annotations to ensure the correctness of the language instructions. Fig.~\ref{fig:attr_to_language_example} (a) presents the examples of the same identity and Fig.~\ref{fig:attr_to_language_example} (b) presents the examples of the transformation with different domains and identities.

\begin{figure*}[p]
  \centering
  \includegraphics[width=\linewidth]
  {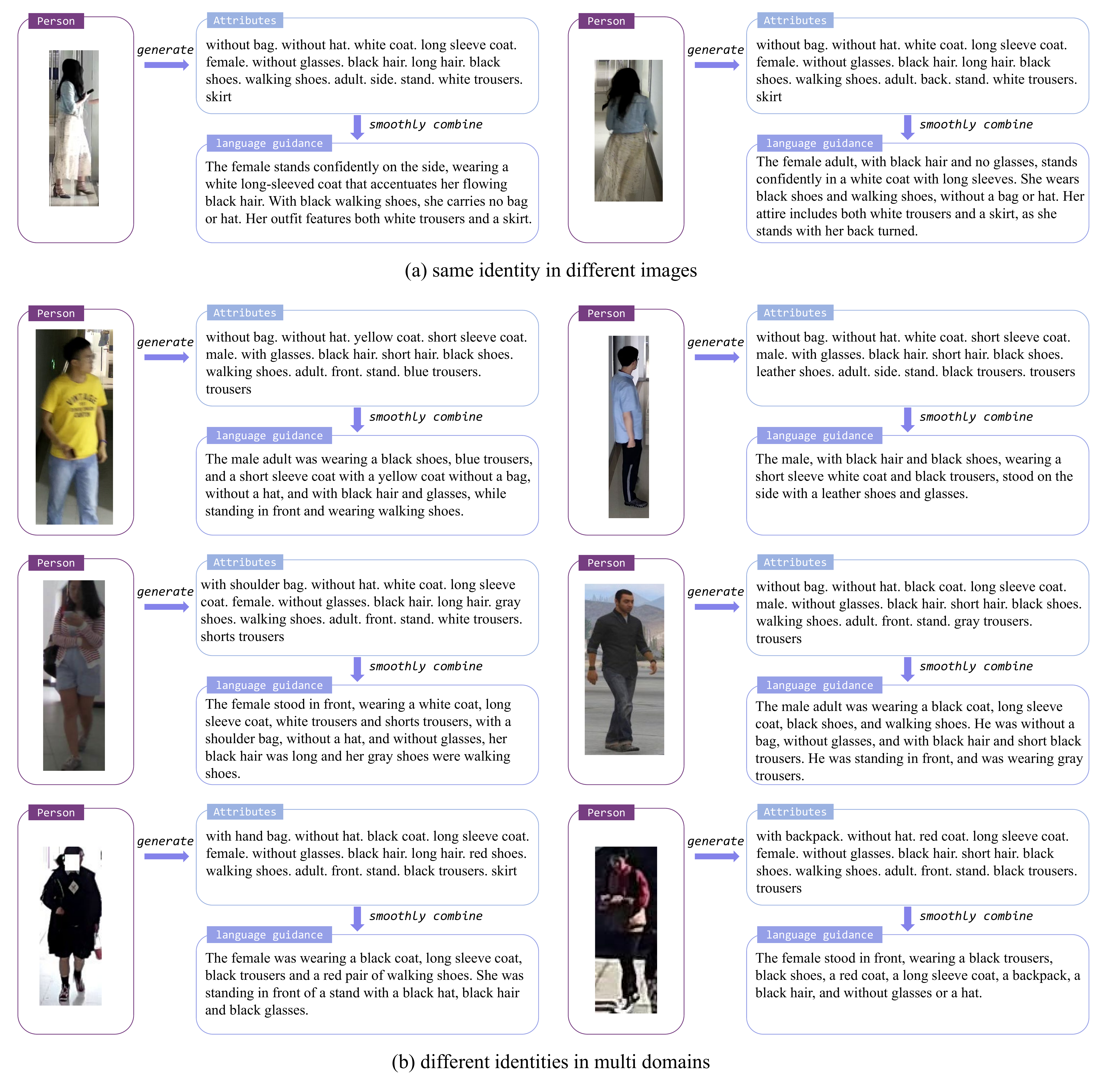}
  \caption{We first generate attributes for a person and then transform attributes into sentences by a large language model. \textbf{(a)} The attributes and language guidance of the same identity. \textbf{(b)} More instances of attributes-to-language transformation with different domains and identities.}
  \label{fig:attr_to_language_example}
\end{figure*}

\vspace{-0.5cm}
\section{Instruction Generation}
In our proposed instruct-ReID task, each identity is further split into query and gallery, where query set consists of query person images and clothes templates, and gallery set consists of target person images. In this section, we provide more training and test examples for Clothes template based clothes-changing ReID and Language-instructed ReID scenarios.
\vspace{-0.3cm}
\subsection{Instructions on traditional ReID, cloth-changing ReID and visual-infrared ReID}
\vspace{-0.2cm}
Following the methods in Instruct-BLIP~\cite{instructblip}, which uses GPT-4 to generate 20 different instructions for traditional ReID (\emph{e.g.}, We use 'do not change clothes' to generate equivalent expressions such as 'maintain consistent clothes' and so on), 20 different instructions for clothes-changing ReID (\emph{e.g.}, We use 'ignore clothes' to generate equivalent expressions such as 'change your clothes' and so on), and 20 different instructions for visual-infrared ReID (\emph{e.g.}, We use 'retrieve cross-modality images' to generate equivalent expressions such as 'fetch images across different modalities' and so on) as shown in Tab.~\ref{tab:instruction_trad_cc} in detail. We randomly choose one from these instructions when training each mini-batch and evaluating every instruction for testing the model.

\begin{table*}[p]\scriptsize
  \centering
  \renewcommand\arraystretch{1.2}
  \caption{Details of instructions for Trad-ReID, CC-ReID, and VI-ReID in OmniReID. We use GPT-4 to generate 20 different expressions for 'do not change clothes‘ as instructions for traditional ReID and similarly, we generated 20 different expressions of 'ignore clothes' for clothes-changing ReID and 20 different expressions of 'retrieve cross-modality images' for visual-infrared ReID.}
  \resizebox{0.96\textwidth}{!}{
    \begin{tabular}{@{}c|c}
    \toprule
    ReID task & instruction representation \\
    \midrule
    {Trad-ReID} & \multicolumn{1}{m{10cm}}{"do not change clothes", "maintain consistent clothes", "keep original clothes", "preserve current clothes", "retain existing clothes", "wear the same clothes", "stick with your clothes", "don't alter your clothes", "no changes to clothes", "unchanged outfit", "clothes remain constant", "no clothing adjustments", "steady clothing choice", "clothing remains unchanged", "consistent clothing selection", "retain your clothing style", "clothing choice remains", "don't swap clothes", "maintain clothing selection", "clothes stay the same"} \\ \hline
    {CC-ReID} & \multicolumn{1}{m{10cm}}{"change your clothes", "swap outfits", "switch attire", "get into a different outfit", "try on something new", "put on fresh clothing", "dress in alternative attire", "alter your outfit", "wear something else", "don a different ensemble", "trade your garments", "shift your wardrobe", "exchange your clothing", "update your attire", "replace your outfit", "clothe yourself differently", "switch your style", "update your look", "put on a new wardrobe", "ignore clothes"} \\ \hline
    {VI-ReID} & \multicolumn{1}{m{10cm}}{"retrieve cross-modality images", "fetch images across different modalities", "collect images from various modalities", "obtain images spanning different modalities", "retrieve images from diverse modalities", "gather images across modalities", "access images across different modalities", "acquire images spanning various modalities", "extract images from different modalities", "retrieve images across multiple modalities", "fetch images from distinct modalities", "collect images across various modalities", "access images from different modalities", "obtain images from diverse modalities", "gather images spanning different modalities", "extract images from various modalities", "retrieve images across varied modalities", "obtain images from distinct modalities", "access images across multiple modalities", "collect images spanning diverse modalities"} \\ 
    \bottomrule
  \end{tabular}}
  \label{tab:instruction_trad_cc}
\end{table*}

\subsection{Text to image ReID}

In T2I-ReID scenario, during the training process, both images and responding description texts are fed into IRM. We adopt a contrastive loss to align the image features and text features. To further enhance the retrieval capability of the model, we employ a classifier to determine whether an inputted image-text pair is positive or negative. Specifically, we retain the original image-text pairs as positives and form negative pairs by matching text features with unrelated image features before inputting them into the attention module. In the inference stage, the query is the describing sentences, and the image features and query features are extracted separately. given a query text feature, we rank all the test gallery image features based on their similarity with the text. We select the top 128 image features and pair them with the query text feature. These pairs are then input into the attention module, further utilizing the matching scores to rank these images. The search is deemed to be successful if top-K images contain any corresponding identity. 

\begin{figure*}[t]
  \centering
  \includegraphics[width=\linewidth]
  {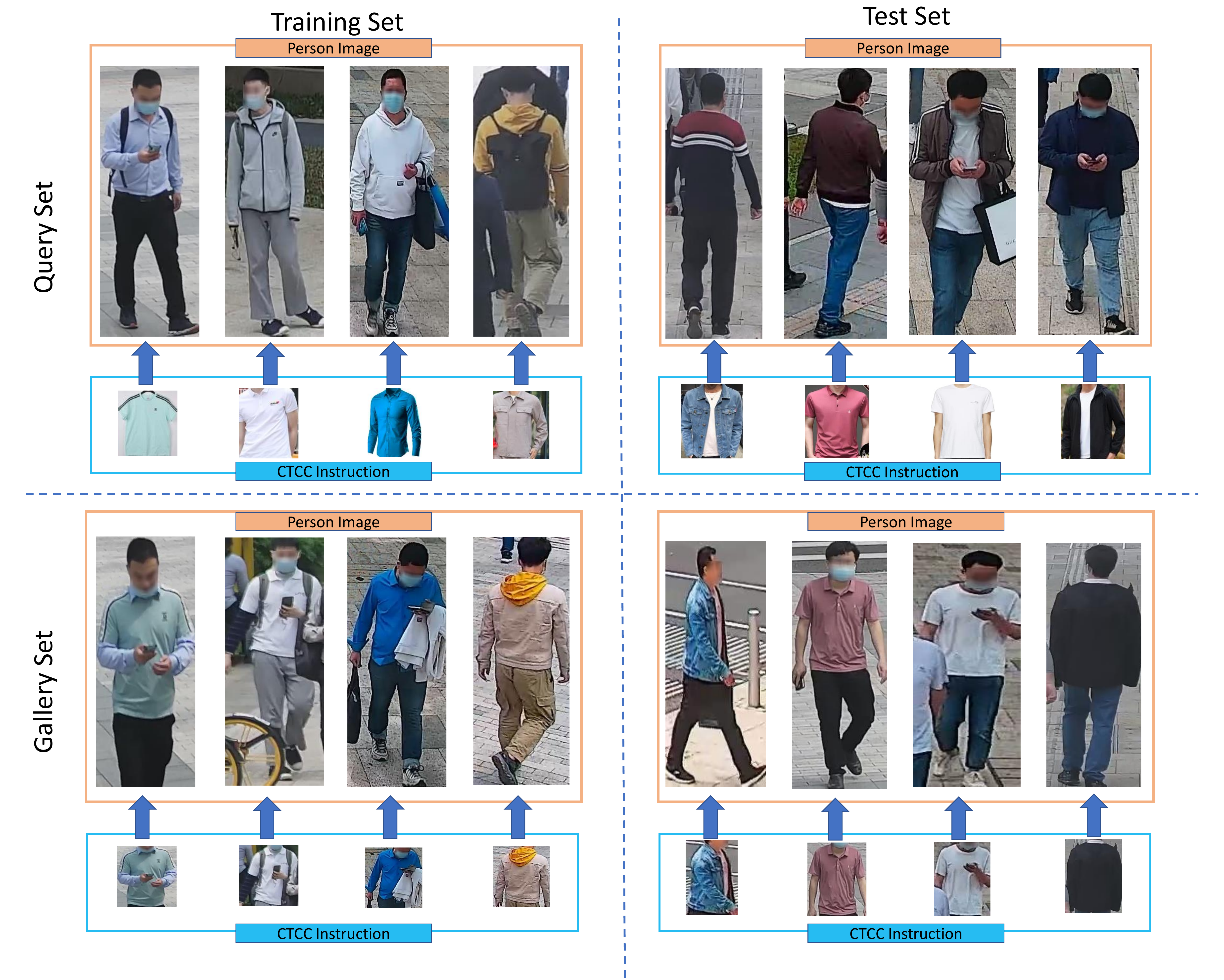}
  \caption{Examples of CTCC-ReID task for training and test. Each identity is further split into query and gallery, where query set consists of query person images and clothes templates as instructions, and gallery set consists of target person images and cropped clothes images as instructions.}
  \label{fig:ctcc_train_test_set_sample}
\end{figure*}
\vspace{-0.2cm}
\subsection{Clothes Template Based Clothes-Changing ReID}
\vspace{-0.2cm}
In CCTC-ReID scenario, as shown in Fig.~\ref{fig:ctcc_train_test_set_sample}, the instruction is a clothes template for a query image, clothes regions cropped by a detector from themselves are treated as instruction for gallery images. During the training process, the biometric feature and clothes feature are extracted from person images and instructions. In the inference stage, the model should retrieve images of the same person wearing the provided clothes.
\vspace{-0.2cm}
\subsection{Language-instructed ReID}
\vspace{-0.2cm}
We provide more examples for LI-ReID as shown in Fig.~\ref{fig:li_train_test_set_sample}. Similar to CTCC-ReID, the instruction for gallery images is several sentences describing pedestrian attributes. We randomly select the description languages from the corresponding person images in gallery and provide to query images as instruction. The model is required to retrieve images of the same person following the provided sentences.

\begin{figure*}
  \centering
  \includegraphics[width=\linewidth]
  {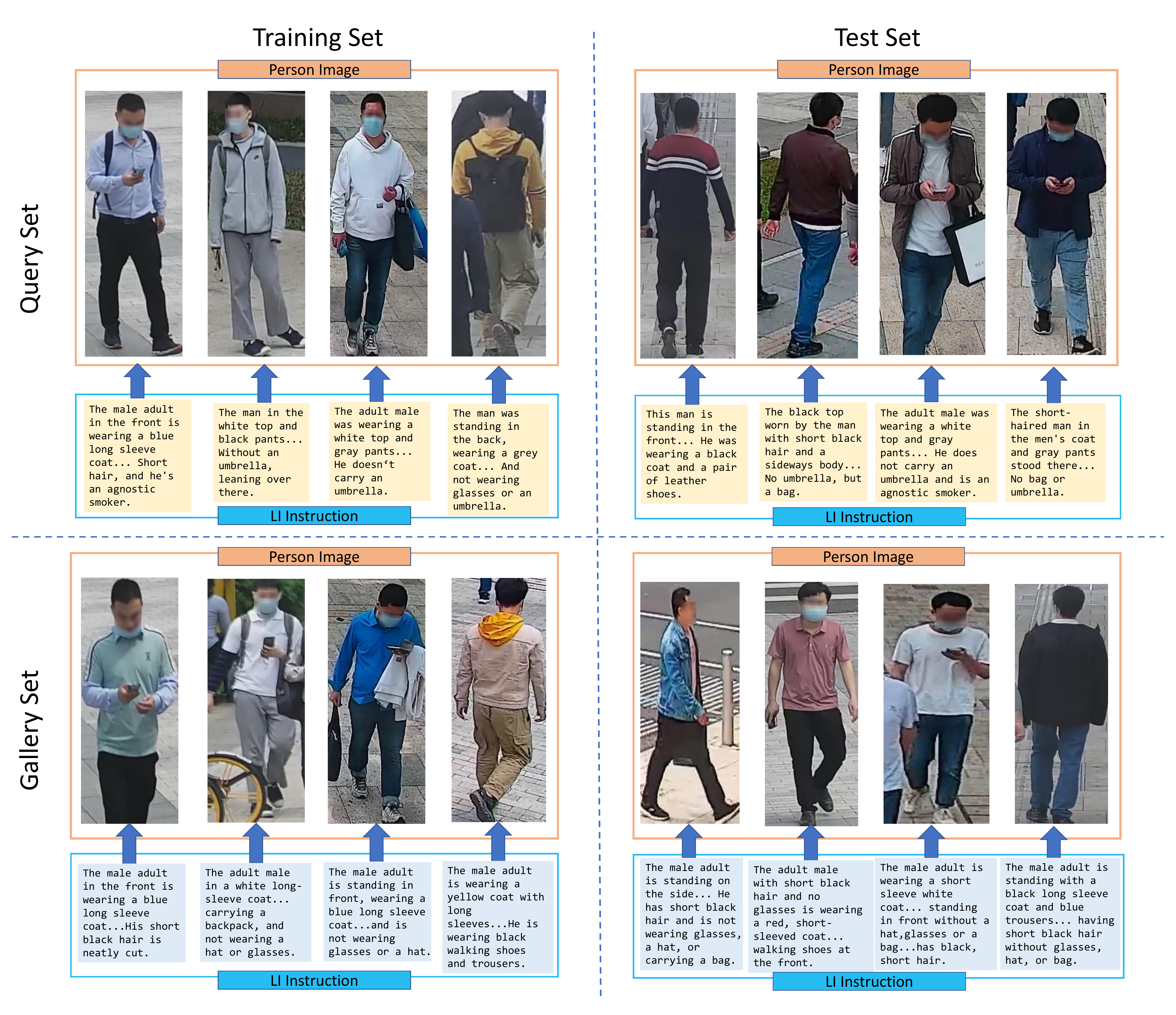}
  \caption{Examples of LI-ReID task for training and test. Each identity is further split into query and gallery, where query set consists of query person images and language instructions (randomly selected from gallery corresponding to the person), and gallery set consists of target person images and description language for themselves as instructions.}
  \label{fig:li_train_test_set_sample}
\end{figure*}

\section{Details of OmniReID}
\vspace{-0.2cm}
In the main text, we briefly introduce the number of images
and number of tasks of OmniReID. For the evaluation of OmniReID, we introduce the evaluation scenario and evaluation protocols. In this section, we present detailed information on the training dataset and evaluation dataset and discuss the ethical issues of these datasets.

\begin{table*}[h]
  \centering
  \renewcommand\arraystretch{1.1}
  \caption{Dataset statistics of OmniReID. We summarize both the training and test set images based on the built benchmark. \dag means the images is the same as the LI-ReID task, we do not calculate the same image twice when calculating the total number of images in Multi-task Learning. \ddag means the LTCC with extra clothes templates for CTCC task.}
    \begin{tabular}{c|cc|cc}
    \toprule
    \multicolumn{1}{c|}{\multirow{2}{*}{\textbf{Single-task learning}}}&\multicolumn{2}{c|}{\textbf{Training}} &\multicolumn{2}{c}{\textbf{Test}} \\
    \cline{2-5} & Dataset & Number of images & Dataset & Number of images \\
    \hline
    Trad-ReID & MSMT17~\cite{wei2018person} & 30,248& MSMT17~\cite{wei2018person} & 93,820 \\
    Trad-ReID & CUHK03~\cite{li2014deepreid} & 7,365& CUHK03~\cite{li2014deepreid} & 6,732 \\
    Trad-ReID & Market1501~\cite{zheng2015person} & 12,936& Market1501~\cite{zheng2015person} & 23,100 \\
    LI-ReID & COCAS+ Real1~\cite{li2022cocas+} & 34,469& COCAS+ Real2~\cite{li2022cocas+} & 14,449 \\
    CC-ReID & PRCC~\cite{yang2019person} & 17,896& PRCC~\cite{yang2019person} & 10,800 \\
    CC-ReID & VC-Clothes~\cite{wan2020person} & 9,449& VC-Clothes~\cite{wan2020person} & 9,611 \\
    CC-ReID & LTCC~\cite{qian2020long} & 9,423& LTCC~\cite{qian2020long} & 7,543 \\
    CTCC-ReID & COCAS+ Real1~\cite{li2022cocas+} & 34,469\dag & COCAS+ Real2~\cite{li2022cocas+} & 14,449\dag \\
    VI-ReID & LLCM~\cite{zhang2023diverse} & 30,921 & LLCM~\cite{zhang2023diverse} &  13,909 \\
    T2I-ReID & CUHK-PEDES~\cite{li2017person} & 28,566 & CUHK-PEDES~\cite{li2017person} & 3,074 \\
    \hline
    \multicolumn{1}{c|}{\multirow{2}{*}{\textbf{Multi-task learning}}}&\multicolumn{2}{c|}{\textbf{Training}} &\multicolumn{2}{c}{\textbf{Test}} \\
    \cline{2-5} & Dataset & Number of images & Dataset & Number of images \\
    \hline
    \multicolumn{1}{c|}{\multirow{12}{*}{OmniReID}} & MSMT17~\cite{wei2018person} & \multicolumn{1}{c|}{\multirow{12}{*}{4,973,044}}& MSMT17~\cite{wei2018person} & 93,820 \\
    & +CUHK03~\cite{li2014deepreid} & & CUHK03~\cite{li2014deepreid} & 6,732 \\
    & +Market1501~\cite{zheng2015person} & & Market1501~\cite{zheng2015person} & 23,100 \\
    & +COCAS+ Real1~\cite{li2022cocas+} & & COCAS+ Real2~\cite{li2022cocas+} & 14,449 \\
    & +PRCC~\cite{yang2019person} & & PRCC~\cite{yang2019person} & 10,800 \\
    & +VC-Clothes~\cite{wan2020person} & & VC-Clothes~\cite{wan2020person} & 9,611 \\
    & +LTCC~\cite{qian2020long} & & LTCC~\cite{qian2020long} & 7,543 \\
    & +COCAS+ Real1~\cite{li2022cocas+} & & COCAS+ Real2~\cite{li2022cocas+} & 14,449\dag \\
    & +LLCM~\cite{zhang2023diverse} & & LLCM~\cite{zhang2023diverse} &  13,909 \\
    & +CUHK-PEDES~\cite{li2017person} & & CUHK-PEDES~\cite{li2017person} & 3,074 \\
    & +LTCC\ddag~\cite{qian2020long} & & - & - \\
    & +SYNTH-PEDES~\cite{zuo2023plip} & & - & - \\
    \bottomrule
    \end{tabular}%
  \label{tab:OmniReID-details}%
\end{table*}%
\vspace{-0.2cm}
\subsection{Dataset Statistics of OmniReID}
\vspace{-0.2cm}
OmniReID collects 12 publicly available datasets of 6 existing ReID tasks, including Traditional ReID (Trad-ReID), Clothes-Changing ReID (CC-ReID), Clothes Template Based Clothes-Changing ReID (CTCC-ReID), Visual-Infrared ReID (VI-ReID), Text-to-Image ReID (T2I-ReID) and Language-Instructed ReID (LI-ReID). As shown in Tab.~\ref{tab:OmniReID-details}, we perform two training scenarios based on the built benchmark \textbf{OmniReID}: 1) Single-task Learning (STL): Every dataset is treated as a single task, which is trained and tested individually. 2) Multi-task Learning (MTL): The model is optimized by joint training of all the ReID tasks with all the training datasets. The trained model is then evaluated on different tasks with various datasets.In Trad-ReID, we utilize the widely-used MSMT17, CUHK03, and Market1501 datasets. For the CC-ReID task, we choose the widely-used PRCC, LTCC, and VC-Clothes datasets, considering the diversity of domains. Regarding the CTCC-ReID task, we label LTCC with clothes image instructions and employ COCAS+ Real1, LTCC for training, while COCAS+ Real2 serves as the test set. For LI-ReID, we annotate the COCAS+ Real1 and COCAS+ Real2 datasets with language instructions for training and test. In VI-ReID, we select a new and challenging low-light cross-modality dataset called LLCM. Finally, for T2I-ReID, we opt for the widely-used CUHK-PEDES and SYNTH-PEDES datasets for training, with CUHK-PEDES serving as the test set. OmniReID forms a total of 4973044 images for training and 183038 images for test, which unites all-purpose person retrieval into one instruct-ReID task.

\subsection{Discussion of Ethical Issues}

The usage of OmniReID might bring several risks, such as privacy, and problematic content. We discuss these risks and their mitigation strategies in this subsection.

First, we conduct a thorough review of each dataset and guarantee that none of the ReID tasks used in our paper are withdrawn. The demographic makeup of the datasets used is not representative of the broader population but these datasets can be used for scientific experimentation.

Second, we adopt the following measures to mitigate potential security risks while adhering to the copyright policies of each dataset:
\begin{itemize}
    \item We will NOT re-release these public datasets but will only provide the download links or webpages when we release the dataset. We do not claim copyright ownership of the original data, and anyone who wants to use the dataset should still be approved by the original assignee.
    \item We will NOT modify these datasets but exclusively provide visual caption annotation files for publicly available datasets. We confirm these annotations do NOT contain identification information.
    \item We obtained explicit permission through email correspondence for annotating the public datasets.
    \item Access to our annotation file links is contingent upon adherence to our scholarly and research-oriented guidelines.
\end{itemize}

Also, we provide an agreement for anyone who wants to use our OmniReID benchmark.

\begin{mdframed}[backgroundcolor=gray!10]

\begin{center}
    \textbf{OmniReID Agreement} 
\end{center} 

This Agreement outlines the terms and conditions governing the use of OmniReID. By signing this agreement, the Recipient agrees to the following terms:

\begin{itemize}
\item The Recipient agrees the OmniReID does not claim Copyrights of Market1501, CUHK03, MSMT17, PRCC, VC-Clothes, LTCC, COCAS+, LLCM, CUHK-PEDES, SYNTH-PEDES and they SHOULD obtain these public datasets from these data providers.

\item The Recipient agrees they must comply with ALL LICENSEs of Market1501, CUHK03, MSMT17, PRCC, VC-Clothes, LTCC, COCAS+, LLCM, CUHK-PEDES, SYNTH-PEDES when they use OmniReID.

\item The Recipient agrees the demographic makeup of OmniReID is not representative of the broader population.
\item The Recipient agrees OmniReID should only be available for non-commercial research purposes. Any other use, in particular any use for commercial purposes, is prohibited.

\item The Recipient agrees not to use the data for any unlawful, unethical, or malicious purposes.

\item The Recipient agrees not to further copy, publish or distribute any portion of the OmniReID.

\item The Recipient agrees [OUR INSTITUTE] reserves the right to terminate the access to the OmniReID at any time.

Name:\\
Organization/Affiliation:\\
Position:\\
Email:\\
Address:\\
Address (Line2):\\
City: Country:\\
Signature: \\ 
Date: \\
\end{itemize}
\end{mdframed}




\end{document}